\documentclass[a4paper,11pt]{article}

\usepackage[T1]{fontenc}
\usepackage[english]{babel}
\usepackage{mathpazo}                 
\usepackage{textcomp}
\usepackage[protrusion=true,expansion=true,tracking=true,final]{microtype}
\usepackage[skip=0.55\baselineskip plus 2pt,indent=0pt]{parskip} 

\usepackage[a4paper,left=27mm,right=27mm,top=26mm,bottom=30mm,
            headsep=8mm,footskip=12mm,heightrounded]{geometry}
\usepackage{etoolbox}
\usepackage{needspace}

\usepackage{mathtools,amssymb,amsthm}
\allowdisplaybreaks[1]

\usepackage{xcolor}
\definecolor{burgundy}{RGB}{128,0,40}       
\definecolor{burgundydark}{RGB}{90,0,28}    
\definecolor{burgundymid}{RGB}{163,38,68}   
\definecolor{burgundysoft}{RGB}{201,120,140} 
\definecolor{burgundywash}{RGB}{246,236,239} 
\definecolor{ink}{RGB}{26,26,29}
\definecolor{rule}{RGB}{128,0,40}
\definecolor{gold}{RGB}{168,130,62}          
\definecolor{goldsoft}{RGB}{212,184,130}     
\usepackage{adjustbox}

\usepackage{graphicx}
\usepackage{tikz}
\usepackage{tikz-cd}
\usetikzlibrary{matrix,positioning,backgrounds,calc,arrows.meta,fit,quotes,shapes.geometric}
\tikzset{
  dvkln/.style   ={line width=0.8pt},
  dvkarr/.style  ={line width=0.8pt,-{Stealth[length=5pt]}},
  dvkdot/.style  ={circle,fill=burgundydark,inner sep=0pt,minimum size=3.4pt},
  dvkbox/.style  ={rounded corners=2.5pt,draw=burgundy,line width=0.8pt,fill=burgundywash,
                   minimum height=1.8em,minimum width=2.3em},
}
\tikzcdset{arrows={line width=0.8pt}, arrow style=tikz, >={Stealth[length=4.5pt]}}
\usepackage{booktabs,array}
\usepackage{enumitem}
\setlist{itemsep=2pt,topsep=4pt,parsep=0pt}

\usepackage{algorithm}
\usepackage[noend]{algpseudocode}
\renewcommand{\algorithmicrequire}{\textbf{Input:}}
\renewcommand{\algorithmicensure}{\textbf{Output:}}
\renewcommand{\algorithmicprocedure}{\textbf{Procedure:}}

\usepackage[labelfont={bf,color=burgundydark},font=small,
            labelsep=period,width=0.94\linewidth]{caption}

\usepackage{titlesec}
\titleformat{\section}[hang]
  {\normalfont\large\bfseries\color{burgundydark}}
  {\color{burgundy}\thesection}{0.75em}{}
  [{\vspace{2.5pt}{\color{burgundy!55}\titlerule[0.7pt]}}]
\titleformat{\subsection}[hang]
  {\normalfont\normalsize\bfseries\color{burgundydark}}
  {\color{burgundymid}\thesubsection}{0.6em}{}
\titleformat{\subsubsection}[hang]
  {\normalfont\normalsize\itshape\color{burgundydark}}
  {\color{burgundymid}\thesubsubsection}{0.6em}{}
\titlespacing*{\section}{0pt}{2.2ex plus .6ex}{1.1ex}
\titlespacing*{\subsection}{0pt}{1.7ex plus .5ex}{0.7ex}
\titlespacing*{\subsubsection}{0pt}{1.3ex plus .4ex}{0.5ex}
\pretocmd{\section}{\Needspace*{7\baselineskip}}{}{\typeout{PATCHFAIL section}}
\pretocmd{\subsection}{\Needspace*{6\baselineskip}}{}{\typeout{PATCHFAIL subsection}}
\pretocmd{\subsubsection}{\Needspace*{5\baselineskip}}{}{\typeout{PATCHFAIL subsubsection}}

\usepackage{fancyhdr}
\renewcommand{\headrule}{\hbox to\headwidth{\color{burgundy!45}\leaders\hrule height 0.4pt\hfill}}
\fancypagestyle{plain}{\fancyhf{}\fancyhead[R]{\footnotesize\thepage}%
  \renewcommand{\headrule}{}}

\numberwithin{equation}{section}
\theoremstyle{definition}

\theoremstyle{remark}

\providecommand{\keywords}[1]{%
  \vspace{0.6em}\noindent{\color{burgundydark}\textbf{Keywords.}}\ #1}

\newcommand\numRowsK{3}
\newcommand\numColsK{3}
\newcommand{\K}[2]{%
    \edef\Kcol##1##2##3{###2}%
    \edef\Krow##1##2##3{\noexpand\Kcol###1}%
    \Krow
        {1 0 1}
        {0 1 0}
        {1 0 1}%
}

\usepackage[numbers,sort]{natbib}     
\usepackage{hyperref}
\usepackage{orcidlink}
\hypersetup{
  pdftitle={Deep-Learning Estimation of Absorbed Dose for Nuclear Medicine Diagnostics},
  pdfauthor={Luciano Melodia},
  pdfsubject={Deep-learning estimation of absorbed dose},
  pdfkeywords={Deep Learning, Dosimetry, Convolutional Neural Networks, Nuclear Medicine},
  bookmarks=true,
  bookmarksnumbered=true,
  colorlinks=true,
  linkcolor=burgundydark,
  citecolor=burgundydark,
  urlcolor=burgundy,
  filecolor=burgundy,
  pdfnewwindow=true
}
\usepackage[nameinlink]{cleveref}
\usepackage{doi}                      

\begin{document}

\title{\color{burgundydark}\bfseries Deep-Learning Estimation of Absorbed Dose\\[2pt]
       for Nuclear Medicine Diagnostics}
\author{Luciano Melodia\,\orcidlink{0000-0002-7584-7287}\\[2pt]
    Chair of Information Science, University of Regensburg\\
    93051 Regensburg, Germany\\
    \texttt{luciano.melodia@stud.uni-regensburg.de}}
\date{31 March 2018\\[3pt]{\normalsize\itshape Revised and re-typeset 23 July 2026}}

\maketitle
\thispagestyle{plain}

\begin{abstract}
\noindent
    In radionuclide therapy with \(^{177}\mathrm{Lu}\), the absorbed-dose distribution can be approximated by convolving the time-integrated activity distribution with a dose voxel kernel for a single tissue type. This approximation is fast but inaccurate: it treats the body as homogeneous and therefore ignores the tissue heterogeneity that governs where energy is deposited. The heterogeneity can be recovered by combining computed tomography and single-photon emission computed tomography with a Monte Carlo transport simulation, at a high computational cost. We investigate whether the map from a local density kernel to the corresponding dose voxel kernel can instead be learned from data by a convolutional neural network, so that density-adapted kernels become available without a full transport calculation for each patient. On held-out patient data, the proposed U-residual architecture reaches a continuous intersection-over-union score of \(0.86\) after \(308\) epochs, with a mean squared error of \(1.24\times 10^{-4}\) on the normalised targets. This generalisation to unseen data indicates that the network approximates, rather than merely memorises, the simulation-based map from density kernels to dose voxel kernels. The network does not replace the underlying transport physics; it approximates the density-to-dose association that a full Monte Carlo transport simulation would otherwise have to supply anew for every patient.
\end{abstract}


\section{Motivation}
\label{motivation}

Nuclear medicine has become central to oncology, both for diagnosis and for therapy. Because cancer incidence rises with age, both the diagnostic image and the therapeutic dose must be made as informative and as accurate as possible. Hybrid imaging modalities, in particular single-photon emission computed tomography combined with computed tomography (SPECT/CT) and positron-emission tomography combined with computed tomography (PET/CT), record both the local tissue density and the spatial distribution of an administered radioactive tracer. Their spatial resolution ranges from millimetres to centimetres, which is fine enough that metabolically active tissue, such as a tumour, stands out clearly against the surrounding healthy tissue in the reconstructed three-dimensional image.

Nuclear medicine treats tumours as well as detecting them. The patient receives a radioactive compound that accumulates in the tumour and irradiates it from within. In therapy with a \(^{177}\mathrm{Lu}\)-labelled radiopharmaceutical, the emitted \(\beta^{-}\) radiation has a short range and deposits its energy locally, so the delivered dose must be controlled precisely; patient-specific dosimetry is therefore required~\cite{botta2011calculation}. To reconstruct the dose one acquires SPECT images at several times after injection, typically at \(4\), \(24\), \(48\), and \(72\) hours, and integrates the resulting activity over time to obtain the number of decays per voxel. The integration of such statistically uncertain image series is itself a substantial problem in its own right, but it lies outside the scope of the present paper, which instead takes the reconstructed decay distribution as its input, so that the difficulties of that time-integration step play no further role in what follows.

We therefore assume throughout that the decay distribution is already known. The next two sections recall the established convolution-based principles of dose estimation, after which the remainder of the paper develops and evaluates a data-driven alternative to them.

\section{Principles of Dose Voxel Calculation}
\label{principles_of_dose_voxel_calculation}

Starting from the spatial distribution of nuclear decays, one seeks to determine the absorbed-dose distribution. A widely used approach in nuclear medicine is based on convolution of the time-integrated activity distribution with a dose voxel kernel. For a source localised pointwise in the central voxel, such a kernel gives the mean dose deposited throughout a finite voxelised neighbourhood of that voxel, for example a cube of \(9^{3}\) voxels centred on the source voxel.

A classical formulation of this approach uses \(S\)-values \cite{petoussi2007patient}. For a source region \(\mathbf r_{U}\) and a target region \(\mathbf r_{T}\),
\(
S(\mathbf r_{T}\leftarrow \mathbf r_{U})
\)
denotes the absorbed dose in the target region per unit time-integrated activity in the source region. If \(\tilde A_{\mathbf r_{U}}\) is the time-integrated activity in the source region, then the mean absorbed dose in the target region is obtained as the linear superposition, over all source regions in the field of view, of their separate contributions to that region:
\begin{align}
    \langle D(\mathbf r_{T}) \rangle          
    =                                         
    \sum_{\mathbf r_{U}}                      
    \tilde A_{\mathbf r_{U}}\,                
    S(\mathbf r_{T}\leftarrow \mathbf r_{U}). 
\end{align}
This is the voxelised form of the Medical Internal Radiation Dose (MIRD) formalism~\cite{bolch1999dosimetry}, in which the mean dose in a target region is written as the linear superposition of the separate contributions made by every source region, each weighted by its own time-integrated activity.

In the present setting, both source and target regions are individual patient voxels. If \(\mathbf v_{T}\) denotes a target voxel and \(\mathbf v_{U}^{(n)}\), \(n=1,\dots,N\), denote the source voxels in the image domain under consideration, then the mean absorbed dose in the target voxel is given by the completely analogous discrete sum taken over all of the individual source voxels
\begin{align}
    \langle D(\mathbf v_{T}) \rangle                
    =                                               
    \sum_{n=1}^{N}                                  
    \tilde A_{\mathbf v_{U}^{(n)}}\,                
    S(\mathbf v_{T}\leftarrow \mathbf v_{U}^{(n)}). 
\end{align}
Here \(\tilde A_{\mathbf v_{U}^{(n)}}\) denotes the time-integrated activity of the \(n\)-th source voxel. Under the assumption of a translation-invariant homogeneous medium, the \(S\)-value depends only on the relative displacement between source and target. In this special case, the dose distribution can be computed as a discrete convolution of the time-integrated activity distribution with a voxel kernel; numerically, this may for example be implemented by fast transform methods \cite{bracewell1984fast}.

The underlying physical kernel is closely related to the dose point kernel, that is, the dose distribution caused by an isotropic point source in a homogeneous medium \cite{sgouros2008three,bodei2008long}. For voxelised applications, the corresponding voxel \(S\)-values are typically determined by Monte Carlo simulation. A radioactive decay in a source voxel is modelled, the emission direction and the interactions of the emitted particles are sampled stochastically, and the energy deposited in the surrounding voxels is accumulated. From the mean deposited energy per nuclear transformation, one obtains, after division by the voxel masses, the absorbed dose per nuclear transformation or, equivalently, per unit time-integrated activity. Monte Carlo methods are standard for this purpose because they model particle transport, scattering, and energy loss in heterogeneous geometries in a physically consistent manner, albeit at a computational cost that is considerable and often prohibitive for routine clinical use~\cite{chiavassa2006validation,reynaert2002parameter,waters2002monte}.

The absorbed dose \(D\) is, by definition, the energy deposited per unit mass. Writing \(\varepsilon(\mathbf r_{T})\) for the deposited energy, \(m(\mathbf r_{T})\) for the voxel mass, \(\rho(\mathbf r_{T})\) for the mass density, and \(V(\mathbf r_{T})\) for the voxel volume, the pointwise definition and its voxel-averaged approximation read
\begin{align}
    D(\mathbf r_{T})                                                          
    =                                                                         
    \frac{\mathrm d\varepsilon}{\mathrm dm}(\mathbf r_{T}),                   
    \quad                                                                     
    \text{and approximately}                                                  
    \quad                                                                     
    D(\mathbf r_{T})                                                          
    =                                                                         
    \frac{\varepsilon(\mathbf r_{T})}{m(\mathbf r_{T})}                       
    =                                                                         
    \frac{\varepsilon(\mathbf r_{T})}{\rho(\mathbf r_{T})\,V(\mathbf r_{T})}.
\end{align}
The SI unit of absorbed dose is the gray, with \(1\,\mathrm{Gy}=1\,\mathrm{J\,kg^{-1}}\); the approximation replaces the local derivative by the mean deposited energy over the finite voxel volume.

This relation makes explicit that the dose distribution depends sensitively on the local mass density and material composition. Biological tissue is markedly inhomogeneous, above all at the interfaces between soft tissue, lung, and bone. A single dose voxel kernel computed for homogeneous soft tissue cannot represent these differences, so the implicit assumption of a spatially uniform medium introduces systematic errors precisely where the anatomy is most heterogeneous, that is, at exactly those clinically important tissue boundaries~\cite{petoussi2007patient}.

\section{Density-Specific Dose Voxel Kernels}
\label{density_specific_dose_voxel_kernels}

Convolution with a dose voxel kernel computed for homogeneous soft tissue is a computationally efficient method for estimating the radiation energy deposited in the body. However, density differences between bone, soft tissue, and air-filled regions are neglected. Since SPECT/CT data are available for each patient, one has, in addition to the activity distribution, spatially resolved information about the anatomical structure and the local tissue density.

To incorporate this information into dosimetry, it is generally not sufficient to convolve the time-integrated activity distribution with a single global dose kernel. Instead, the spatially varying material distribution must be taken into account. To this end, the CT image is first registered to and resampled onto the voxel grid of the SPECT data. The CT values are then converted, by means of an appropriate calibration, into material or mass-density information. From these data, one can extract local density kernels, or more generally local transport environments, for which corresponding dose voxel kernels are then computed by Monte Carlo simulation. In a fully patient-specific calculation, both the time-integrated activity distribution and the density distribution derived from CT enter the dose computation.

The underlying workflow may be summarised as follows:
\begin{enumerate}[noitemsep]
    \item acquisition of a whole-body CT of the patient;
    \item alignment of the CT data with the image grid of the functional imaging modality;
    \item conversion of CT values into local density or material distributions;
    \item Monte Carlo calculation of the local dose deposition;
    \item reconstruction and temporal integration of the activity distribution;
    \item computation of the dose distribution taking both activity and density into account.
\end{enumerate}

The direct Monte Carlo approach is physically very accurate because it explicitly accounts for particle transport, scattering, energy loss, and material interfaces. Its essential drawback, however, is its high computational cost. By contrast, convolution with a single homogeneous dose voxel kernel is fast, but systematically inaccurate in anatomically heterogeneous regions. This motivates an intermediate approach: one that exploits the anatomical information contained in the CT scan without demanding, for every patient, a completely new and computationally expensive Monte Carlo transport calculation, carried out at full spatial resolution and then repeated entirely from scratch for every individual patient.

The approach pursued here is to learn, in a data-driven manner, the map from local density distributions to the corresponding dose voxel kernels. More precisely, the goal is to construct a model that predicts, from a given local density distribution, a density-adapted dose kernel that can subsequently be used for fast dose computation. In this way, one does not replace the underlying physical dose formation itself, but rather approximates the association between density structure and local energy transport that is otherwise obtained from Monte Carlo simulation. The aim is therefore to combine the accuracy of density-dependent dosimetry with the computational efficiency of convolution-based methods, which is the central trade-off that motivates the present study and shapes the network design described below.

We therefore turn to deep learning. The next section reviews the neural architectures relevant to this task and then motivates the model chosen for the experiment. Throughout, bold lowercase letters denote column vectors and bold uppercase letters denote matrices, and every activation function is understood to act componentwise on its argument.

\section{Background: Deep Learning}
\label{background_deep_learning}

Deep learning is a class of machine-learning methods in which multilayer neural networks learn, from data, successive representations at different levels of abstraction. A central feature of such models is that useful features need not be engineered entirely by hand, but can instead be learned from data during training. This automatic learning of representations has driven substantial advances across many application areas, among them natural-language processing, computer vision, image analysis, and medical imaging, the last of these being the concrete clinical domain from which the dosimetry problem studied in this paper is drawn.

Deep neural networks are usually trained by gradient-based optimisation. One minimises a loss functional that measures the discrepancy between the model prediction and the desired output. The relevant gradients are typically computed by backpropagation. In this way the parameters are adjusted so that the desired output representation is built up, layer by layer, from a given input representation, each successive layer transforming the representation produced by the layer beneath it into one better adapted to the target.

Within machine learning, one distinguishes in particular between supervised, unsupervised, and reinforcement learning. In supervised learning, one learns a map from inputs to target values from labelled data; typical tasks are classification and regression. In unsupervised learning, by contrast, the emphasis lies on structure discovery, feature extraction, or dimensionality reduction without prescribed target values. In the present setting one faces a supervised regression problem, since the goal is to predict a continuous, spatially resolved target quantity from a spatially resolved input representation of the very same spatial size.

Different network architectures are particularly well suited to different data types. Recurrent neural networks are designed for sequential or time-dependent data because they incorporate previous states into the current computation. Convolutional networks, by contrast, are particularly well suited to image-like and spatially structured data because they capture local patterns by means of shared filters and thereby exploit spatial structure efficiently. In medical image analysis, where the data are inherently spatial, convolution-based architectures are therefore among the most important and most widely used model classes.

The relevant ingredients for the present work are the principles of convolution, feature extraction across successive layers, dimensionality reduction by pooling, and gradient-based optimisation. Since both input and target are spatially structured fields, a convolution-based architecture is the natural choice, and the aim is to learn a map from a spatial representation of the local density to a spatial representation of the corresponding dose response.

\section{Supervised Learning with Neural Networks}
\label{supervised_learning_with_neural_networks}

Supervised learning is among the best studied and most widely used paradigms in machine learning. One is given a data set consisting of inputs together with corresponding target values. The goal is to learn from these examples a map that sends new inputs to their target quantities as accurately as possible. Depending on the task, this is either a classification problem or a regression problem. In classification, one assigns to an input one of finitely many classes, whereas in regression one predicts a continuous target quantity. In the present setting, one is dealing with a supervised regression problem, since one seeks to determine a continuous spatial output quantity from a spatial input representation \cite{lecun2015deep}.

A neural network is a sequence of layers, each of which composes an affine map with a nonlinear activation function. A single layer with input \(\mathbf x\in\mathbb R^{d}\), weight matrix \(\mathbf W\in\mathbb R^{m\times d}\), and bias vector \(\mathbf b\in\mathbb R^{m}\) produces the \(m\)-dimensional output vector \(\mathbf y\in\mathbb R^{m}\) given by
\begin{align}
    \mathbf y = \varphi(\mathbf W \mathbf x + \mathbf b),
    \label{eq:layer_activity}
\end{align}
where the activation function \(\varphi\) is applied componentwise. Typical choices for \(\varphi\) are the sigmoid function, the hyperbolic tangent, and, in modern networks, one of the variants of the rectified linear unit \cite{lecun2015deep}.
During training, the parameters \(\mathbf W\) and \(\mathbf b\) are adjusted so that the predictions of the network agree with the target values as closely as possible. To this end, one defines a loss function that measures the discrepancy between prediction and target. In regression tasks, the mean squared error is often used. In classification tasks, one often uses the cross-entropy. Minimisation of this loss function is usually carried out by gradient-based methods, and the required derivatives through all layers are computed by backpropagation, an efficient recursive organisation of the chain rule that is described in detail below \cite{lecun2015deep}.

For orientation, it is useful to begin with a single artificial neuron. In the simplest case of binary classification, a neuron decides on the basis of the sign of an affine function. For an input pattern \(\mathbf x\) and parameters \(\mathbf w\) and \(b\), the binary decision rule of the neuron is
\begin{align}
    y = \operatorname{sign}(\mathbf w^{\top}\mathbf x + b). 
\end{align}
The equation \(\mathbf w^{\top}\mathbf x + b = 0\) defines a hyperplane in \(\mathbb R^{d}\), which divides the input space into two half-spaces. It follows that a single such neuron can solve only linearly separable classification problems. Nonlinearly separable problems, for example of XOR type, generally require multilayer architectures with nonlinear activation functions, which is exactly the additional expressive power that stacking several nonlinear layers into a deep network provides.

Historically, elementary learning rules play an important role for simple linear models. A classical example is the \(\delta\)-rule, which adjusts each weight in proportion both to its share of the output error and to the corresponding input activation, and takes the form
\begin{align}
    w_{j n}^{\mathrm{new}}                   
    =                                        
    w_{j n}^{\mathrm{old}}                   
    +                                        
    \eta\,\delta_{j}^{(\mu)}\,x_{n}^{(\mu)}, 
    \label{eq:delta_rule}                    
\end{align}
where \(\eta > 0\) is the learning rate, \(x_{n}^{(\mu)}\) the \(n\)-th component of the \(\mu\)-th training example, and \(\delta_{j}^{(\mu)}\) an error-dependent correction term for the \(j\)-th output component. In linear models, this rule yields, in its Widrow--Hoff or LMS interpretation, a gradient method for minimising a quadratic error \cite{lippmann1987introduction}. An exact solution in finitely many steps is not guaranteed in general. Rather, convergence depends on the model class, on the choice of learning rate, and on the statistical properties of the training data, and only under further assumptions \cite{lippmann1987introduction}.

The strength of deep neural networks lies in the fact that several layers are composed, thereby allowing highly nonlinear maps to be learned. Recurrent neural networks are particularly suitable for sequential data because they retain information about previous states. Convolutional networks, by contrast, are particularly suitable for spatially structured data because they detect local patterns by means of shared filters and thereby exploit the spatial organisation of the data \cite{lecun2015deep}. Since in the present problem both input and output are spatial fields, convolution-based architectures form the natural model class, and it is precisely such an architecture that we develop and evaluate in the remainder of this work.

Accordingly, we view neural networks here not as biological models but as parametrised classes of functions, to be fitted from sample data so as to realise a map between the two spatial representations that jointly define the dosimetric problem here.

\section{Learning by Gradient Descent}
\label{learning_with_gradient_descent}

In gradient descent, the parameters of a neural network are adjusted iteratively so as to minimise a prescribed loss function. The optimisation process is typically stopped once a termination criterion is met, for example a maximal number of epochs, a sufficiently small change in the loss function, or the absence of further improvement on a validation set. In realistic applications, one should not expect the loss function to attain the value \(0\) exactly.

The loss function measures the discrepancy between the target values and the model predictions. Which loss is appropriate depends on the task. For regression problems, the mean squared error is a common choice. For classification problems, one usually works with logarithmic loss functions, in particular with the cross-entropy. The advantage of such logarithmic losses is not that they generically produce fewer local minima, but rather that they are natural for probabilistic models and, in combination with suitable output functions, especially Softmax, yield computationally and statistically favourable gradients.

Let
\(
\mathcal D=\{(\mathbf x^{(\mu)},\mathbf t^{(\mu)})\mid \mu=1,\dots,M\}
\)
be a data set with inputs \(\mathbf x^{(\mu)}\) and target values \(\mathbf t^{(\mu)}\), and let
\(
\hat{\mathbf y}^{(\mu)}=\mathcal H(\mathbf x^{(\mu)}\mid \mathbf W)
\)
denote the prediction of the network with parameters \(\mathbf W\) on the input of the \(\mu\)-th example. The loss functions that recur most often in this setting are the mean squared error for regression and the cross-entropy for classification, which we now record.

For a regression task with continuous targets, the mean squared error averages the squared Euclidean distance between prediction and target over the entire data set and reads
\begin{align}
    \mathcal L_{\mathrm{MSE}}(\mathbf W)      
    =                                         
    \frac{1}{M}                               
    \sum_{\mu=1}^{M}                          
    \bigl\|                                   
    \hat{\mathbf y}^{(\mu)}-\mathbf t^{(\mu)} 
    \bigr\|_{2}^{2}.                          
\end{align}

For binary classification with target values \(t^{(\mu)}\in\{0,1\}\) and predictions \(\hat y^{(\mu)}\in(0,1)\):
\begin{align}
    \mathcal L_{\mathrm{BCE}}(\mathbf W)                    
    =                                                       
    -                                                       
    \frac{1}{M}                                             
    \sum_{\mu=1}^{M}                                        
    \Bigl(                                                  
    t^{(\mu)}\log \hat y^{(\mu)}                            
    +                                                       
    \bigl(1-t^{(\mu)}\bigr)\log\bigl(1-\hat y^{(\mu)}\bigr) 
    \Bigr).                                                 
\end{align}
The derivative of this averaged binary loss with respect to a single prediction \(\hat y^{(\mu)}\) is then
\begin{align}
    \frac{\partial \mathcal L_{\mathrm{BCE}}}{\partial \hat y^{(\mu)}}
    =
    \frac{1}{M}\,
    \frac{\hat y^{(\mu)}-t^{(\mu)}}{\hat y^{(\mu)}\bigl(1-\hat y^{(\mu)}\bigr)}.
\end{align}

For multiclass classification with target vectors \(\mathbf t^{(\mu)}\) and predicted probability vectors \(\hat{\mathbf y}^{(\mu)}\), the multiclass cross-entropy loss, averaged over the training examples, is defined by
\begin{align}
    \mathcal L_{\mathrm{CE}}(\mathbf W)   
    =                                     
    -                                     
    \frac{1}{M}                           
    \sum_{\mu=1}^{M}                      
    \sum_{k=1}^{K}                        
    t_{k}^{(\mu)}\log \hat y_{k}^{(\mu)},
\end{align}
which, whenever the target vectors are one-hot encoded, reduces to the negative logarithm of the probability that the network assigns to the single correct class of the example.

The Kullback--Leibler divergence between two discrete probability distributions \(p\) and \(q\) is
\begin{align}
    D_{\mathrm{KL}}(p\|q)                  
    =                                      
    \sum_{k} p_{k}\log\frac{p_{k}}{q_{k}}. 
\end{align}
When \(p\) is the fixed target distribution, minimising \(D_{\mathrm{KL}}(p\|q)\) over \(q\) differs from minimising the cross-entropy only by the entropy of \(p\), a quantity that does not depend on \(q\); the two objectives therefore have exactly the same minimiser and the same gradient with respect to \(q\).

\subsection{Forward Propagation}
\label{forward_propagation}

Forward propagation is the step-by-step computation of the network output from a given input. One propagates an input vector through the successive layers of the network. Each layer first applies an affine map to its input and then an activation function. For the \(\ell\)-th layer, with its own weight matrix and bias vector, this computation takes the form
\begin{align}
    \mathbf h^{(\ell)}                                        
    =                                                         
    \varphi^{(\ell)}                                          
    \bigl(                                                    
    \mathbf W^{(\ell)}\mathbf h^{(\ell-1)}+\mathbf b^{(\ell)} 
    \bigr),                                                   
\end{align}
where \(\mathbf h^{(0)}=\mathbf x\) is the input. This process is continued up to the output layer.

For classification tasks, it is convenient to interpret the output as a probability distribution over the classes. To this end, one often uses the Softmax function in the last layer. For a real score vector \(\mathbf z=(z_{1},\dots,z_{K})\in\mathbb R^{K}\), the Softmax output is defined componentwise by
\begin{align}
    \operatorname{softmax}(\mathbf z)_{k}       
    =                                           
    \frac{e^{z_{k}}}{\sum_{j=1}^{K} e^{z_{j}}}, 
    \qquad k=1,\dots,K.                         
\end{align}
Each component is nonnegative, and one has
\(
\sum_{k=1}^{K}\operatorname{softmax}(\mathbf z)_{k}=1.
\)
The output can therefore be interpreted directly as a vector of class probabilities.
In what follows,
\(
\mathcal H(\mathbf x\mid \mathbf W)
\)
denotes the output produced by the network with parameters \(\mathbf W\) on input \(\mathbf x\). For the example shown in Fig.~\ref{forward}, one should note that the diagram displays five input components and five output components. Accordingly, we consider below an input vector of length \(5\).

\begin{figure}[t]
    \centering
  \adjustbox{max width=\linewidth}{\footnotesize\color{burgundydark}
\begin{tikzcd}[column sep=7.5em, row sep=3.2em]
    {}                                                                  & {}                                                               & {}                                                              & {}                                                               & {}                                                                  \\
    \bullet \arrow[u, "\mathbf{y}_1" description]                       & \bullet \arrow[u, "\mathbf{y}_2" description]                    & \bullet \arrow[u, "\mathbf{y}_3" description]                   & \bullet \arrow[u, "\mathbf{y}_4" description]                    & \bullet \arrow[u, "\mathbf{y}_5" description]                       \\
    \bullet \arrow[u] \arrow[ru] \arrow[rru] \arrow[rrru] \arrow[rrrru] & \bullet \arrow[lu] \arrow[u] \arrow[ru] \arrow[rru] \arrow[rrru] & \bullet \arrow[llu] \arrow[lu] \arrow[u] \arrow[ru] \arrow[rru] & \bullet \arrow[lllu] \arrow[llu] \arrow[lu] \arrow[u] \arrow[ru] & \bullet \arrow[llllu] \arrow[lllu] \arrow[llu] \arrow[lu] \arrow[u] \\
    {} \arrow[u, "\mathbf{x}_1" description]                            & {} \arrow[u, "\mathbf{x}_2" description]                         & {} \arrow[u, "\mathbf{x}_3" description]                        & {} \arrow[u, "\mathbf{x}_4" description]                         & {} \arrow[u, "\mathbf{x}_5" description]
\end{tikzcd}}
    \caption{Schematic depiction of forward propagation in a fully connected network, where each unit of a layer is joined to all units of the next layer.}
    \label{forward}
\end{figure}

\subsection{Backpropagation}
\label{backpropagation}

Backpropagation is the standard method for the efficient computation of the gradients of a loss function in multilayer neural networks. It is based on a systematic application of the chain rule and yields, for all weights and bias parameters, the partial derivatives of the loss function. Combined with a gradient-based update rule, these derivatives let all parameters be adjusted iteratively, at a total cost comparable to that of a single additional forward pass through the network, which is what makes gradient training of large models feasible at all~\cite{rumelhart1986learning,lippmann1987introduction}.

We first consider a fully connected network with \(M\) layers. Fix a training example \((\mathbf x,\mathbf t)\) and set \(\mathbf a^{(0)}=\mathbf x\). For each layer \(m=1,\dots,M\), the pre-activation \(\mathbf z^{(m)}\) and the activation \(\mathbf a^{(m)}\) of every layer are then defined by the following forward-propagation recursion
\begin{align}
    \mathbf z^{(m)}=\mathbf W^{(m)}\mathbf a^{(m-1)}+\mathbf b^{(m)},
    \qquad
    \mathbf a^{(m)}=\varphi^{(m)}\bigl(\mathbf z^{(m)}\bigr),
\end{align}
in which \(\mathbf W^{(m)}\) is the weight matrix of the \(m\)-th layer, \(\mathbf b^{(m)}\) its bias vector, and \(\varphi^{(m)}\) the activation function applied componentwise; the output of the network is then read off as \(\hat{\mathbf y}=\mathbf a^{(M)}\), the activation produced by the final layer of the forward pass.

For the derivation of the backpropagation algorithm, we first consider the quadratic error incurred by the network on a single fixed training example, written explicitly as
\begin{align}
    \mathcal L(\mathbf x,\mathbf t\mid \mathbf W,\mathbf b) 
    =                                                       
    \frac{1}{2}                                             
    \bigl\|                                                 
    \hat{\mathbf y}-\mathbf t                               
    \bigr\|_{2}^{2}                                         
    =                                                       
    \frac{1}{2}                                             
    \sum_{i}                                                
    \bigl(a_{i}^{(M)}-t_{i}\bigr)^{2}.                      
    \label{eq:bp_loss_single}                               
\end{align}
The factor \(1/2\) serves only to simplify the derivatives. For a data set
\(
\mathcal D=\{(\mathbf x^{(\mu)},\mathbf t^{(\mu)})\mid \mu=1,\dots,N\},
\)
the total loss of the network is then obtained by summing, or equivalently averaging, this single-example error over all of the training examples in the data set.

To compute the gradients, one introduces for each layer the error signal
\begin{align}
    \boldsymbol{\delta}^{(m)}                             
    :=                                                    
    \frac{\partial \mathcal L}{\partial \mathbf z^{(m)}}. 
\end{align}
For the output layer, one obtains the output error signal directly from its definition, since applying the chain rule to the quadratic loss function yields immediately
\begin{align}
    \boldsymbol{\delta}^{(M)}                            
    =                                                    
    \frac{\partial \mathcal L}{\partial \mathbf a^{(M)}} 
    \odot                                                
    \varphi^{(M)\prime}\bigl(\mathbf z^{(M)}\bigr)          
    =                                                    
    \bigl(\mathbf a^{(M)}-\mathbf t\bigr)                
    \odot                                                
    \varphi^{(M)\prime}\bigl(\mathbf z^{(M)}\bigr),         
    \label{eq:delta_output}                              
\end{align}
where \(\odot\) denotes componentwise multiplication.

For a hidden layer \(m\in\{1,\dots,M-1\}\), the chain rule gives the recursion
\begin{align}
    \boldsymbol{\delta}^{(m)}                    
    =                                            
    \bigl(\mathbf W^{(m+1)}\bigr)^{\top}         
    \boldsymbol{\delta}^{(m+1)}                  
    \odot                                        
    \varphi^{(m)\prime}\bigl(\mathbf z^{(m)}\bigr). 
    \label{eq:delta_hidden}                      
\end{align}
This is the actual backward propagation of the error: the error signal of the next higher layer is propagated back via the transpose of the weight matrix and is then multiplied by the derivative of the local activation function, evaluated at the corresponding pre-activation.

From the error signals one obtains the gradients with respect to the weights and bias parameters. For each layer \(m\), the gradients take the outer-product and vector form
\begin{align}
    \frac{\partial \mathcal L}{\partial \mathbf W^{(m)}} 
    =                                                    
    \boldsymbol{\delta}^{(m)}                            
    \bigl(\mathbf a^{(m-1)}\bigr)^{\top},                
    \qquad                                               
    \frac{\partial \mathcal L}{\partial \mathbf b^{(m)}} 
    =                                                    
    \boldsymbol{\delta}^{(m)}.                           
    \label{eq:gradients_bp}                              
\end{align}
Written componentwise for a single weight, the first of these two formulas reads
\begin{align}
    \frac{\partial \mathcal L}{\partial w_{ij}^{(m)}} 
    =                                                 
    \delta_{i}^{(m)} a_{j}^{(m-1)}.                   
\end{align}

A gradient-descent step with learning rate \(\eta > 0\) therefore has the form
\begin{align}
    \mathbf W^{(m)}_{\mathrm{new}}
      & = 
    \mathbf W^{(m)}_{\mathrm{old}}
    -
    \eta
    \frac{\partial \mathcal L}{\partial \mathbf W^{(m)}},
    \\
    \mathbf b^{(m)}_{\mathrm{new}}
      & = 
    \mathbf b^{(m)}_{\mathrm{old}}
    -
    \eta
    \frac{\partial \mathcal L}{\partial \mathbf b^{(m)}}.
    \label{eq:gradient_step_bp}
\end{align}
This is the precise form of the rule often referred to, in elementary expositions, as the \(\delta\)-rule.

In practice, the loss function is usually not evaluated on each individual training example, but rather on small subsets of the data set. Such a subset is called a mini-batch. An epoch is a complete pass through the whole training set. The random permutation of the training examples between epochs and optimisation on mini-batches lead to stochastic or mini-batch gradient descent. A general convergence theorem for non-convex deep networks in full generality is not available. In practice, stopping criteria are chosen, for example, on the basis of the number of epochs, the development of the validation error, or the size of the gradients.

Summarising, the following procedure results for a network with \(M\) layers:
\begin{enumerate}[noitemsep]
    \item Initialise the weights and biases with small random values.
    \item Choose a training example or a mini-batch.
    \item Compute \(\mathbf z^{(m)}\) and \(\mathbf a^{(m)}\) successively by forward propagation for all \(m=1,\dots,M\).
    \item Compute the error signal of the output layer by \eqref{eq:delta_output}.
    \item Compute the error signals of the hidden layers recursively by \eqref{eq:delta_hidden}.
    \item Determine the gradients by \eqref{eq:gradients_bp}.
    \item Update the parameters by \eqref{eq:gradient_step_bp}.
    \item Repeat steps \(2\) through \(7\) until a stopping criterion is satisfied.
\end{enumerate}

This presentation is mathematically equivalent to classical backpropagation, but is notationally consistent and immediately usable for multilayer networks. In particular one must keep notationally separate, throughout the whole derivation, the pre-activations \(\mathbf z^{(m)}\), the activations \(\mathbf a^{(m)}\), the target values \(\mathbf t\), and the error signals \(\boldsymbol{\delta}^{(m)}\), since these four families of quantities play formally distinct roles at every stage of the derivation.

\begin{figure}
    \centering
  \adjustbox{max width=\linewidth}{\footnotesize\color{burgundydark}\begin{tikzcd}
                                                                                          & {}                                                                                      & {}                                                                                      &                                                                                           &  &                                                                                           &                                                                                         & {} \arrow[d]                                                                            &                                                                                           \\
                                                                                          & \bullet \arrow[u, "\mathbf{y}_1" description]                                           & \bullet \arrow[u, "\mathbf{y}_2" description]                                           &                                                                                           &  &                                                                                           & \bullet                                                                                 & \bullet \arrow[d, no head]                                                              &                                                                                           \\
\bullet \arrow[ru, no head] \arrow[rru, no head]                                          & \bullet \arrow[u, no head] \arrow[ru, no head]                                          & \bullet \arrow[lu, no head] \arrow[u, no head]                                          & \bullet \arrow[lu, no head] \arrow[llu, no head]                                          &  & \bullet                                                                                   & \bullet                                                                                 & \bullet \arrow[lld, no head] \arrow[ld, no head] \arrow[d, no head] \arrow[rd, no head] & \bullet                                                                                   \\
\bullet \arrow[u, no head] \arrow[ru, no head] \arrow[rru, no head] \arrow[rrru, no head] & \bullet \arrow[lu, no head] \arrow[u, no head] \arrow[ru, no head] \arrow[rru, no head] & \bullet \arrow[ru, no head] \arrow[u, no head] \arrow[lu, no head] \arrow[llu, no head] & \bullet \arrow[lllu, no head] \arrow[llu, no head] \arrow[lu, no head] \arrow[u, no head] &  & \bullet \arrow[d, no head] \arrow[rd, no head] \arrow[rrd, no head] \arrow[rrrd, no head] & \bullet \arrow[rd, no head] \arrow[ld, no head] \arrow[d, no head] \arrow[rrd, no head] & \bullet \arrow[lld, no head] \arrow[ld, no head] \arrow[d, no head] \arrow[rd, no head] & \bullet \arrow[llld, no head] \arrow[lld, no head] \arrow[ld, no head] \arrow[d, no head] \\
\bullet \arrow[u, no head] \arrow[ru, no head] \arrow[rru, no head] \arrow[rrru, no head] & \bullet \arrow[lu, no head] \arrow[u, no head] \arrow[ru, no head] \arrow[rru, no head] & \bullet \arrow[llu, no head] \arrow[lu, no head] \arrow[u, no head] \arrow[ru, no head] & \bullet \arrow[lllu, no head] \arrow[llu, no head] \arrow[lu, no head] \arrow[u, no head] &  & \bullet                                                                                   & \bullet                                                                                 & \bullet                                                                                 & \bullet                                                                                   \\
{} \arrow[u, "\mathbf{x}_1" description]                                                  & {} \arrow[u, "\mathbf{x}_2" description]                                                & {} \arrow[u, "\mathbf{x}_3" description]                                                & {} \arrow[u, "\mathbf{x}_4" description]                                                  &  &                                                                                           &                                                                                         &                                                                                         &                                                                                          
\end{tikzcd}}
    \caption{Schematic depiction of forward propagation on the left and backward propagation of the error on the right, reusing the same connections.}
\end{figure}

\section{Development and Variants of Recurrent Networks}
\label{development_and_variants_of_recurrent_neural_networks}

Recurrent neural networks do not form a single architecture, but rather a class of models characterised by the fact that the state of the network at a given time depends on earlier states. In contrast to purely feedforward networks, they therefore process not only the current input but also internal state information from previous steps. It is therefore natural, mathematically, to regard a recurrent network as a discrete-time dynamical system whose state evolves under the successive inputs and carries information from one step to the next~\cite{lukovsevivcius2009reservoir,lecun2015deep}.

This architecture is particularly suitable for sequential or time-dependent data, such as language, text, measurement series, or other sequences of observations. The essential advantage of recurrent models is that they can store information about previous inputs in a hidden state and incorporate it into subsequent processing. In this sense, they possess a finite, data-dependent short-term memory. The precise pattern of recurrent connections varies from model to model; in the typical case, however, the hidden state produced at one time step re-enters the computation at the following time step, providing a short-term memory~\cite{lecun2015deep,lukovsevivcius2009reservoir}.

Historically, Hopfield networks and Boltzmann machines belong to the best-known recurrent models \cite{hopfield1982neural,ackley1985learning}. In addition, recurrent networks became established for the modelling of time-dependent data, in particular in the form of simple recurrent networks, Elman networks, and later LSTM architectures \cite{elman1991distributed,hochreiter1997long}. These should be distinguished from residual networks, whose skip connections constitute a different, non-recurrent architectural idea \cite{he2016identity}. Likewise, deep belief networks are not recurrent networks, but deeply layered generative models.

\subsection{Simple Recurrent Neural Networks}
\label{simple_recurrent_neural_networks}

One of the classical forms of recurrent networks is the simple recurrent network due to Elman \cite{elman1991distributed}. In this architecture, one takes into account not only the current input \(\mathbf x_{t}\), but also the hidden state \(\mathbf h_{t-1}\) from the previous time step. For a time step \(t\), the model has the form
\begin{align}
    \mathbf h_{t}                 
    =                             
    \varphi\bigl(                    
    \mathbf W_{xh}\mathbf x_{t}   
    +                             
    \mathbf W_{hh}\mathbf h_{t-1} 
    +                             
    \mathbf b_{h}                 
    \bigr),                       
    \label{eq:elman_hidden}       
\end{align}
where \(\mathbf h_{t}\) denotes the hidden state, \(\mathbf W_{xh}\) the input weights, \(\mathbf W_{hh}\) the recurrent weights, and \(\mathbf b_{h}\) the bias vector. The scalar or vector output may then be defined, for example,
\begin{align}
    \mathbf y_{t}               
    =                           
    \psi\bigl(                  
    \mathbf W_{hy}\mathbf h_{t} 
    +                           
    \mathbf b_{y}               
    \bigr).                     
    \label{eq:elman_output}     
\end{align}

The context units used in older expositions are not an independent physical memory mechanism, but rather an illustrative interpretation of the recurrent state: the hidden state of one time step is stored and re-injected at the next step. Formally, this is already completely described by \eqref{eq:elman_hidden}. Such networks are trained not by ordinary backpropagation on a single-layer graph, but by backpropagation through time, in which the recurrent computation is unfolded into a feedforward graph along the time axis and then differentiated~\cite{rumelhart1986learning,lippmann1987introduction}.

\begin{figure}[t]
    \centering
  \adjustbox{max width=\linewidth}{\footnotesize\color{burgundydark}\begin{tikzcd}
                                                 &                                                &                                              & {}                                                             & {}                                                              & {}                                                                 &                                             \\
                                                 &                                                &                                              & \bullet \arrow[u, "\mathbf{y}_1" description]                  & \bullet \arrow[u, "\mathbf{y}_2" description]                   & \bullet \arrow[u, "\mathbf{y}_3" description]                      &                                             \\
\color{gold}\bullet_+ \arrow[rrrrr, bend right] & \color{gold}\bullet_+ \arrow[rrr, bend right] & \color{gold}\bullet_+ \arrow[r, bend right] & \bullet \arrow[u] \arrow[ru] \arrow[rru] \arrow[l, bend right] & \bullet \arrow[lu] \arrow[u] \arrow[ru] \arrow[lll, bend right] & \bullet \arrow[lu] \arrow[llu] \arrow[u] \arrow[lllll, bend right] &                                             \\
                                                 &                                                & \bullet \arrow[ru] \arrow[rru] \arrow[rrru]  & \bullet \arrow[u] \arrow[ru] \arrow[rru]                       & \bullet \arrow[lu] \arrow[u] \arrow[ru]                         & \bullet \arrow[llu] \arrow[lu] \arrow[u]                           & \bullet \arrow[lllu] \arrow[llu] \arrow[lu] \\
                                                 &                                                & {} \arrow[u, "\mathbf{x}_1" description]     & {} \arrow[u, "\mathbf{x}_2" description]                       & {} \arrow[u, "\mathbf{x}_3" description]                        & {} \arrow[u, "\mathbf{x}_4" description]                           & {} \arrow[u, "\mathbf{x}_5" description]   
\end{tikzcd}}
    \caption{Schematic representation of an Elman-type simple recurrent network with input, hidden, and output layers, in which the hidden state is fed back into the hidden layer and forms the network's recurrent memory.}
\end{figure}

\subsection{Learning Sequences with Recurrent Neural Networks}
\label{learning_of_sequences_with_recurrent_neural_networks}

Recurrent neural networks are used in particular for the modelling of time series. A time series is a sequence of observations
\(
\mathbf x_{1},\dots,\mathbf x_{T},
\)
whose entries are measured at successive time points. Typical examples are biosignals, speech signals, financial data, or other medical measurement series. A recurrent network processes such a sequence one element at a time and, at each step, updates a hidden state that summarises the relevant information seen so far in the past.

In the simplest form, one obtains, for each \(t=1,\dots,T\), the coupled recursion
\begin{align}
    \mathbf h_{t}
      & = 
    \varphi\bigl(
    \mathbf W_{xh}\mathbf x_{t}
    +
    \mathbf W_{hh}\mathbf h_{t-1}
    +
    \mathbf b_{h}
    \bigr),
    \\
    \mathbf y_{t}
      & = 
    \psi\bigl(
    \mathbf W_{hy}\mathbf h_{t}
    +
    \mathbf b_{y}
    \bigr).
\end{align}
If one wishes to represent the entire sequence by a single final state, then the entire output sequence may instead be represented by a single vector formed only at the last time point:
\begin{align}
    \mathbf y                   
    =                           
    \psi\bigl(                  
    \mathbf W_{hy}\mathbf h_{T} 
    +                           
    \mathbf b_{y}               
    \bigr).                     
\end{align}

The weights \(\mathbf W_{xh}\), \(\mathbf W_{hh}\), and \(\mathbf W_{hy}\) are identical across all time steps. Precisely this weight sharing makes recurrent networks suitable for sequences of variable length. The input sequences therefore need not necessarily be padded with zeros to a common length, although this is sometimes still done in concrete implementations or for mini-batch processing for practical reasons. Conceptually, however, the natural strength of recurrent networks lies in the direct processing of sequences of variable and a priori unbounded length \cite{lecun2015deep}.

\begin{figure}[t]
    \centering
  \adjustbox{max width=\linewidth}{\footnotesize\color{burgundydark}\begin{tikzcd}
    &&&   &  &   &  & \bullet^{\mathbf{y}}    \\
{\color{burgundy}\bullet^{\mathbf{h}_1}} \arrow[r, "u" description] & {\color{burgundy}\bullet^{\mathbf{h}_2}} \arrow[r, "u" description] & {\color{burgundy}\bullet^{\mathbf{h}_3}} \arrow[r, "u" description] & \color{burgundy}\bullet^{\mathbf{h}_4} \arrow[r, "u" description] & \color{burgundy}\bullet^{\mathbf{h}_5}  & \cdots & \color{burgundy}\bullet^{\mathbf{h}_{T-1}} \arrow[r, "u" description] & \color{burgundy}\bullet^{\mathbf{h}_{T}} \arrow[u, "v" description] \\
\bullet \arrow[u, "w" description]& \bullet \arrow[u, "w" description]& \bullet \arrow[u, "w" description]& \bullet \arrow[u, "w" description]   & \bullet \arrow[u, "w" description]  & \cdots & \bullet \arrow[u, "w" description]  & \bullet \arrow[u, "w" description]\\
{} \arrow[u, "\mathbf{x}_1" description]    & {} \arrow[u, "\mathbf{x}_2" description]    & {} \arrow[u, "\mathbf{x}_3" description]    & {} \arrow[u, "\mathbf{x}_4" description]  & {} \arrow[u, "\mathbf{x}_5" description] &   & {} \arrow[u, "\mathbf{x}_{T-1}" description]  & {} \arrow[u, "\mathbf{x}_T" description]   
\end{tikzcd}}
    \caption{Schematic representation of a recurrent network for processing an input sequence. The hidden state is propagated from one time step to the next.}
\end{figure}

\section{LSTM Architectures in Recurrent Neural Networks}
\label{long_short_term_memory_in_recurrent_neural_networks}

Training simple recurrent networks over long time horizons is difficult because, during backpropagation through time, gradients typically become either very small or very large. This problem is known as the vanishing-gradient and exploding-gradient phenomenon \cite{hochreiter1997long}. It arises because norms are multiplied repeatedly in long products of Jacobian matrices, which can lead to exponential damping or exponential growth. The difficulty therefore does not primarily stem from an uneven information density in individual input dimensions, but rather from the recursive structure of the derivatives across many time steps.

LSTM architectures were introduced specifically to mitigate this difficulty \cite{hochreiter1997long}. The central idea is to maintain, alongside the hidden state \(\mathbf h_{t}\), a cell state \(\mathbf c_{t}\) along which information can be propagated over longer time intervals under gate control. In one standard and widely used form, the LSTM gate and state variables are governed by the coupled equations
\begin{equation}
    \begin{aligned}
        \mathbf f_{t}
          & = 
        \sigma\bigl(
        \mathbf W_{xf}\mathbf x_{t}
        +
        \mathbf W_{hf}\mathbf h_{t-1}
        +
        \mathbf b_{f}
        \bigr),
        \\
        \mathbf i_{t}
          & = 
        \sigma\bigl(
        \mathbf W_{xi}\mathbf x_{t}
        +
        \mathbf W_{hi}\mathbf h_{t-1}
        +
        \mathbf b_{i}
        \bigr),
        \\
        \tilde{\mathbf c}_{t}
          & = 
        \tanh\bigl(
        \mathbf W_{xc}\mathbf x_{t}
        +
        \mathbf W_{hc}\mathbf h_{t-1}
        +
        \mathbf b_{c}
        \bigr),
        \\
        \mathbf c_{t}
          & = 
        \mathbf f_{t}\odot \mathbf c_{t-1}
        +
        \mathbf i_{t}\odot \tilde{\mathbf c}_{t},
        \\
        \mathbf o_{t}
          & = 
        \sigma\bigl(
        \mathbf W_{xo}\mathbf x_{t}
        +
        \mathbf W_{ho}\mathbf h_{t-1}
        +
        \mathbf b_{o}
        \bigr),
        \\
        \mathbf h_{t}
          & = 
        \mathbf o_{t}\odot \tanh(\mathbf c_{t}).
        \label{eq:lstm_hidden_state}
    \end{aligned}
\end{equation}
Here \(\sigma\) denotes the sigmoid function and \(\odot\) the componentwise product. The forget gate \(\mathbf f_{t}\) controls which portion of the old cell state is retained. The input gate \(\mathbf i_{t}\) determines how strongly new information is written into the cell state. The output gate \(\mathbf o_{t}\) regulates which portion of the current cell state becomes visible in the outgoing hidden state.

The stabilisation of learning is based on the fact that the cell state has a largely additive dynamics. As a result, under suitable gate values, the error signal can be transported across many time steps without necessarily vanishing or exploding immediately \cite{hochreiter1997long}. This largely additive dynamics of the cell state is what allows LSTM networks to model long-range temporal dependencies substantially better than simple recurrent networks can.

The gates are not free scalars, but vector-valued, data-dependent functions of the current input and the previous hidden state. This adaptive control fundamentally distinguishes the LSTM from a simple recurrent network. Variants of the basic model arise through modifications of the gate structure, through additional peephole connections, through bidirectional processing in both time directions, or through coupling with convolutional operations for spatially and temporally structured data such as video and volumetric image sequences.

\begin{figure}
    \centering
  \adjustbox{max width=\linewidth}{\footnotesize\color{burgundydark}
\begin{tikzpicture}[>=Stealth, x=1.06cm, y=1.06cm,
  op/.style   ={circle, draw=burgundydark, line width=0.8pt, minimum size=1.75em, inner sep=0pt, fill=white},
  gate/.style ={rounded corners=3pt, draw=burgundy, line width=0.8pt, fill=burgundywash,
                minimum height=1.9em, minimum width=2.4em},
  cand/.style ={rounded corners=3pt, draw=gold, line width=0.8pt, fill=gold!16!white,
                minimum height=1.9em, minimum width=2.6em},
  fun/.style  ={rounded corners=3pt, draw=burgundydark, line width=0.8pt, fill=white,
                minimum height=1.9em, minimum width=2.6em},
  cw/.style   ={draw=burgundydark, line width=0.8pt, ->},
  hw/.style   ={draw=burgundymid, line width=0.8pt, ->},
  lab/.style  ={font=\scriptsize, inner sep=1.5pt}]

  \node        (cin)  at (-0.8,3.0) {$\mathbf c_{t-1}$};
  \node[op]    (mf)   at ( 1.5,3.0) {$\odot$};
  \node[op]    (add)  at ( 4.3,3.0) {$\oplus$};
  \coordinate  (tap)  at ( 6.5,3.0);
  \node        (cout) at ( 9.6,3.0) {$\mathbf c_{t}$};
  \draw[cw] (cin)--(mf);  \draw[cw] (mf)--(add);  \draw[cw] (add)--(cout);
  \fill[burgundydark] (tap) circle (1.6pt);

  \node[fun] (tc) at (6.5,1.95) {$\tanh$};
  \node[op]  (mo) at (7.9,1.95) {$\odot$};
  \draw[cw] (tap)--(tc);
  \draw[cw] (tc)--(mo);
  \draw[hw] (mo)--(9.6,1.95) node[right,inner sep=2pt]{$\mathbf h_{t}$};

  \node[gate] (f) at (1.5,1.0) {$\sigma$};
  \node[gate] (i) at (3.4,1.0) {$\sigma$};
  \node[cand] (g) at (5.0,1.0) {$\tanh$};
  \node[op]   (mi) at (4.2,2.0) {$\odot$};
  \node[gate] (o) at (7.9,1.0) {$\sigma$};

  \draw[cw] (f)--(mf) node[lab,burgundy,pos=0.5,right]{$\mathbf f_t$};
  \draw[cw] (i)--(mi) node[lab,burgundy,pos=0.55,left=1pt]{$\mathbf i_t$};
  \draw[cw] (g)--(mi) node[lab,gold,pos=0.55,right=1pt]{$\tilde{\mathbf c}_t$};
  \draw[cw] (mi)--(add);
  \draw[cw] (o)--(mo) node[lab,burgundy,pos=0.5,left=1pt]{$\mathbf o_t$};

  \node[lab] (inlab) at (-0.8,-0.5) {$[\mathbf h_{t-1},\,\mathbf x_{t}]$};
  \draw[hw,-] (inlab.east)--(8.6,-0.5);
  \foreach \g in {f,i,g,o} { \draw[hw] (\g.south|-inlab.south)--(\g.south); }
\end{tikzpicture}%
}
    \caption{Schematic representation of an LSTM cell in a recurrent network. The cell state transports information along the time axis, and the gates adaptively regulate how this information is written, retained, and read.}
\end{figure}

\section{Convolutional Neural Networks}
\label{convolutional_neural_networks}

Convolutional neural networks are among the most important deep-learning architectures for spatially structured data. They have had a major impact on image classification, object detection, and semantic segmentation, and they achieve very strong performance on many standardised image data sets \cite{lecun2015deep,simonyan2014very,redmon2016you,ren2015faster,long2015fully}. Their particular advantage lies in the fact that they capture local spatial structure with a manageable number of parameters and are therefore also suitable for image-to-image mappings in which one seeks to predict a spatially structured output image from an input image of the same spatial dimensions \cite{lecun2015deep}.

Mathematically, convolution in analysis is a bilinear operation that maps two functions to a third function. For two integrable functions \(f,g\colon \mathbb R^{d}\to\mathbb R\), their continuous convolution is the function whose value at each point \(x\) is defined by the convolution integral
\begin{align}
    (f*g)(x)                                                
    =                                                       
    \int_{\mathbb R^{d}} f(\tau)\,g(x-\tau)\,\mathrm d\tau. 
\end{align}
In convolutional neural networks, however, one works with discrete data. For an image, or more generally a two-dimensional signal \(\mathbf X\), and a filter \(\mathbf K\), one uses in practice usually not the mathematically exact discrete convolution, but rather the discrete cross-correlation. For a filter of size \((2r+1)\times(2s+1)\), this cross-correlation at output position \((x,y)\) is given by
\begin{align}
    \mathbf T_{x,y}                 
    =                               
    \sum_{n=-r}^{r}\sum_{m=-s}^{s}  
    \mathbf K_{n,m}\,               
    \mathbf X_{x+n,y+m}.            
    \label{eq:cross_correlation_2d} 
\end{align}
Here \(\mathbf X\) denotes the input matrix, \(\mathbf K\) the learnable filter, and \(\mathbf T\) the resulting feature map. In the deep-learning literature, this operation is nevertheless usually referred to as convolution, although strictly speaking it is a cross-correlation \cite{lecun2015deep}. For the mathematical structure of the architectures, however, this conventional distinction is immaterial here.

In a convolutional neural network, such a filter is not applied at a single position only, but translated across the entire input signal. This gives rise to local receptive fields and to a substantial reduction in the number of parameters compared with fully connected layers. While a fully connected layer for an image of size \(150\times 150\) would already require \(22500\) input weights per neuron, a \(3\times 3\) filter uses only \(9\) weights per input channel, independently of the image size. This reuse of the same small set of weights at every spatial position is the central structural advantage of convolution-based models over fully connected ones~\cite{lecun2015deep}.

Another essential point is the translation equivariance of the convolution operation. If the input signal is shifted spatially, then, under idealised boundary handling, the resulting feature map is shifted in the same way. Convolution alone therefore does not produce translation invariance, but rather translation equivariance. A certain invariance with respect to small shifts arises only through additional architectural elements such as pooling, striding, or global aggregation \cite{lecun2015deep}. This distinction between equivariance and invariance is important here, since the two are frequently conflated in informal presentations of convolutional networks.

\begin{figure}
    \centering
  \adjustbox{max width=\linewidth}{\footnotesize\color{burgundydark}
\begin{tikzpicture}
    \tikzset{%
        parenthesized/.style={left delimiter = (, right delimiter = )},
        node distance = 7mu,
        every matrix/.append style={row sep=-1.5pt, column sep=-1pt},
    }

    \matrix[matrix of math nodes, parenthesized] (I) {
        0 & 1 & 1 & 1 & 0 & 0 & 0 \\
        0 & 0 & 1 & 1 & 1 & 0 & 0 \\
        0 & 0 & 0 & 1 & 1 & 1 & 0 \\
        0 & 0 & 0 & 1 & 1 & 0 & 0 \\
        0 & 0 & 1 & 1 & 0 & 0 & 0 \\
        0 & 1 & 1 & 0 & 0 & 0 & 0 \\
        1 & 1 & 0 & 0 & 0 & 0 & 0 \\
    };

    \node (*) [right = of I] {${}*{}$};

    \newcommand\Kmatrix{}
    \foreach \row in {1, ..., 3} {
        \gdef \sep {}
        \foreach \col in {1, ..., 3} {%
            \xdef \Kmatrix {\unexpanded\expandafter{\Kmatrix}\unexpanded\expandafter{\sep}\noexpand \K{\row}{\col}}
            \gdef \sep { \& }
        }
        \xdef \Kmatrix {\unexpanded\expandafter{\Kmatrix}\noexpand\\}
    }
    \matrix[matrix of math nodes, parenthesized, ampersand replacement=\&] (K) [right = of *] {
        \Kmatrix
    };

    \node (=) [right = of K] {${}={}$};

    \matrix[matrix of math nodes, parenthesized] (I*K) [right = of {=}] {
        1 & 4 & 3 & 4 & 1 \\
        1 & 2 & 4 & 3 & 3 \\
        1 & 2 & 3 & 4 & 1 \\
        1 & 3 & 3 & 1 & 1 \\
        3 & 3 & 1 & 1 & 0 \\
    };

    \newcommand\rowResult{1}
    \newcommand\colResult{4}

    \begin{scope}[on background layer]
        \newcommand{\padding}{2pt}
        \coordinate (Is-nw) at ([xshift=-\padding, yshift=+\padding] I-\rowResult-\colResult.north west);
        \coordinate (Is-se) at ([xshift=+\padding, yshift=-\padding] I-\the\numexpr\rowResult+\numRowsK-1\relax-\the\numexpr\colResult+\numColsK-1\relax.south east);
        \coordinate (Is-sw) at (Is-nw |- Is-se);
        \coordinate (Is-ne) at (Is-se |- Is-nw);

        \filldraw[burgundy, fill opacity=.12] (Is-nw) rectangle (Is-se);
        \filldraw[burgundymid, fill opacity=.16] (I*K-\rowResult-\colResult.north west) rectangle (I*K-\rowResult-\colResult.south east);

        \draw[gold, densely dotted]
            (Is-nw) -- (K.north west)
            (Is-se) -- (K.south east)
            (Is-sw) -- (K.south west)
            (Is-ne) -- (K.north east)
        ;
        \draw[burgundymid, densely dotted]
            (I*K-\rowResult-\colResult.north west) -- (K.north west)
            (I*K-\rowResult-\colResult.south east) -- (K.south east)
            (I*K-\rowResult-\colResult.south west) -- (K.south west)
            (I*K-\rowResult-\colResult.north east) -- (K.north east)
        ;

        \draw[gold, fill=gold!14!white] (K.north west) rectangle (K.south east);
    \end{scope}

    \tikzset{node distance=0em}
    \node[below=2pt of I] (I-label) {$\mathbf{X}$};
    \node at (K |- I-label)     {$\mathbf{K}$};
    \node at (I*K |- I-label)   {$\mathbf{T}^{(x,y)}$};
\end{tikzpicture}%
}
    \caption{Schematic representation of the application of a local filter of size \(3\times 3\) to a two-dimensional input signal. The weighted sum over the current receptive field is written into the corresponding entry of the output matrix.}
\end{figure}

\subsection{Weight Sharing and Translation Equivariance}
\label{translation_invariance_of_convolution}

The true source of the efficiency of convolutional neural networks is the combination of local coupling and weight sharing. A filter is used with the same parameters at all spatial positions. Thus the network does not learn a separate system of feature detectors for each position, but rather the same local detector for all positions. This encodes the assumption that local patterns, such as edges, textures, or transitions, are relevant in a comparable way at different locations of the image, so that a detector learned at one position is useful everywhere \cite{lecun2015deep}.

Weight sharing may be formulated algebraically as follows. There are not independent weights for each spatial position, but rather a single shared filter tensor \(\mathbf K\), whose entries in \eqref{eq:cross_correlation_2d} are reused at every position. Accordingly, in backpropagation one does not update several independent weights, but instead sums the gradients arising from all spatial applications of the same filter. For a parameter \(\theta\) of a filter, the derivative of the loss function is therefore
\begin{align}
    \frac{\partial \mathcal L}{\partial \theta}          
    =                                                    
    \sum_{(x,y)}                                         
    \frac{\partial \mathcal L}{\partial \mathbf T_{x,y}} 
    \frac{\partial \mathbf T_{x,y}}{\partial \theta}.    
\end{align}
This summation over all spatial positions is the precise mathematical form of the shared filter update. It is therefore not correct to begin with several distinct weights and then identify them afterwards by averaging. Rather, there exists from the outset only a single shared parameter set, which is used at all positions and is updated by a single summed gradient \cite{lecun2015deep}.

In classical architectures, convolution layers are often followed by nonlinear activations and by spatially reducing operations. Historically, local contrast normalisation and related normalisation techniques were also used. In modern architectures, however, batch normalisation, instance normalisation, or similar methods usually take over the role of stabilising the scale of activations. For this reason local contrast normalisation is today of mainly historical rather than of foundational or systematic interest in convolutional architectures~\cite{lecun2015deep}.

\subsection{Pooling}
\label{pooling}

Pooling methods serve to compress feature maps spatially. To this end, one moves a local window of fixed size across a feature matrix and applies a prescribed aggregation operation within that window. Common variants are max-pooling and average pooling. Further variants, such as stochastic pooling or pyramid pooling, have also been studied \cite{boureau2011ask,boureau2010theoretical,zeiler2013stochastic,lee2016generalizing}.

For a feature matrix \(\mathbf X\) and a pooling window \(P_{x,y}\), the two most widely used variants are max-pooling and average pooling, which respectively return the largest entry and the arithmetic mean of the feature-map entries that fall inside the current pooling window at that position:
\begin{align}
    \mathbf Y^{\max}_{x,y}
      & =
    \max_{(i,j)\in P_{x,y}}
    \mathbf X_{i,j},
    & \mathbf Y^{\mathrm{avg}}_{x,y}
      & =
    \frac{1}{|P_{x,y}|}
    \sum_{(i,j)\in P_{x,y}}
    \mathbf X_{i,j}.
\end{align}
In addition, one fixes a stride, the number of pixels by which the pooling window is shifted at each step; the stride controls both the spatial resolution of the resulting output matrix and the overall degree of dimensional reduction performed by the pooling operation.

Pooling reduces the spatial resolution and at the same time enlarges the effective receptive field of later layers. This allows the network to form coarser and more robust feature representations. At the same time, however, spatial detail is lost. Whether pooling is useful therefore depends strongly on the task. For classification problems, this compression is often advantageous, whereas for dense prediction tasks, such as segmentation or image-to-image regression, it must be used with caution, since there precise spatial localisation is essential \cite{lecun2015deep,long2015fully}. In the present setting, where one seeks to predict a spatially highly resolved dose response, this architectural trade-off between resolution and abstraction is particularly important.

\section{Stochastic Gradient Descent}
\label{stochastic_gradient_descent}

Stochastic gradient descent is one of the basic optimisation methods used in the training of neural networks. Its aim is to minimise an empirical or expected loss function without having to compute, at each iteration step, the full gradient over the entire data set. Instead, the gradient is estimated from single randomly chosen training examples or from small subsets of the data set. Especially for large data sets and high-dimensional parameter spaces, this is computationally much cheaper than a full gradient descent step over the entire data set.

Consider a data set
\(
\mathcal D=\{z_{1},\dots,z_{N}\},
\
z_{i}=(\mathbf x_{i},\mathbf t_{i}),
\)
a parametrised loss function \(\ell(\mathbf w;z)\) for a single example \(z\), and the empirical objective function
\(
F(\mathbf w)
=
\frac{1}{N}\sum_{i=1}^{N}\ell(\mathbf w;z_{i}).
\)
Full gradient descent updates the parameters \(\mathbf w_{t}\) against the averaged gradient, according to
\begin{align}
    \mathbf w_{t+1}                                                               
    =                                                                             
    \mathbf w_{t}                                                                 
    -                                                                             
    \eta_{t}\nabla F(\mathbf w_{t})                                               
    =                                                                             
    \mathbf w_{t}                                                                 
    -                                                                             
    \eta_{t}\frac{1}{N}\sum_{i=1}^{N}\nabla_{\mathbf w}\ell(\mathbf w_{t};z_{i}), 
    \label{eq:gd_correct}                                                         
\end{align}
where \(\eta_{t} > 0\) denotes the learning rate in step \(t\). Stochastic gradient descent replaces the full gradient by a random estimator. For a single randomly selected example \(z_{i_{t}}\), the update is
\begin{align}
    \mathbf w_{t+1}                                          
    =                                                        
    \mathbf w_{t}                                            
    -                                                        
    \eta_{t}\nabla_{\mathbf w}\ell(\mathbf w_{t};z_{i_{t}}). 
    \label{eq:sgd_correct}                                   
\end{align}
Often one uses, instead of a single example, a mini-batch \(B_{t}\subseteq\{1,\dots,N\}\) with \(|B_{t}|=b\), in which case the averaged parameter update over the mini-batch becomes
\begin{align}
    \mathbf w_{t+1}                                                                  
    =                                                                                
    \mathbf w_{t}                                                                    
    -                                                                                
    \eta_{t}\frac{1}{b}\sum_{i\in B_{t}}\nabla_{\mathbf w}\ell(\mathbf w_{t};z_{i}). 
    \label{eq:minibatch_sgd}                                                         
\end{align}
This formula contains full gradient descent as the case \(b=N\) and pure stochastic gradient descent as the case \(b=1\), with intermediate sizes interpolating between the extremes.

The essential advantage of stochastic gradient descent is that each iteration step is cheap and may also be carried out online as soon as new data arrive. The price for this is the variance of the gradient estimator. Hence the parameter sequence does not, in general, follow exactly the direction of the full gradient, but rather a noisy approximation to it. In practice, however, this noise is often acceptable and may even help the method move through flat regions or unfavourable local structures of the loss landscape. Theoretical statements about convergence rates depend on the assumptions imposed on the objective function, on the variance of the estimator, and on the choice of the learning-rate schedule. For suitable learning-rate schedules, for example \(\eta_{t}\sim t^{-1}\) under additional regularity assumptions, one obtains classical convergence statements in convex settings. For deep non-convex networks, by contrast, there is no comparably general theory with global guarantees.

The expressions in \eqref{eq:gd_correct} and \eqref{eq:sgd_correct} at the same time correct a common sign error in informal presentations. Since one seeks to minimise, the gradient must be subtracted rather than added. Likewise, stochastic gradient descent does not arise as a limiting case of full gradient descent; on the contrary, it is full gradient descent that appears as the special case of the mini-batch rule in which the mini-batch is taken to be the entire training data set at once.

\renewcommand{\algorithmicrequire}{\textbf{Input:}}
\renewcommand{\algorithmicensure}{\textbf{Output:}}
\renewcommand{\algorithmicprocedure}{\textbf{Procedure:}}

\subsection{Adam as an Adaptive First-Order Method}
\label{sgd_with_adaptive_lower_order_momentum}

For the training of deep networks, adaptive variants of stochastic gradient descent are often used. A particularly widespread method is \texttt{Adam}. The name stands for \emph{adaptive moment estimation}. The method combines ideas from the momentum method with adaptive coordinatewise learning rates, as they also occur in \texttt{AdaGrad} and \texttt{RMSProp}. More precisely, \texttt{Adam} maintains exponentially weighted running estimates of the first moment and of the second uncentred moment of the stochastic gradient, and rescales each step accordingly.

Let \(\mathbf\Theta\) be the parameter vector and let \(\mathcal L_{t}(\mathbf\Theta)\) be the loss function evaluated on the mini-batch used in step \(t\). Then the stochastic gradient is
\(
\mathbf g_{t}
=
\nabla_{\mathbf\Theta}\mathcal L_{t}(\mathbf\Theta_{t-1}).
\)
From the stochastic gradient \(\mathbf g_t\), the method maintains two exponentially weighted moving averages, the first- and second-moment estimates \(\mathbf m_t\) and \(\mathbf v_t\); it corrects their bias and advances the parameters by an adaptively scaled step, according to the coupled parameter-update rules
\begin{align}
    \mathbf m_{t} &= \beta_{1}\mathbf m_{t-1}+(1-\beta_{1})\mathbf g_{t},
      & \mathbf v_{t} &= \beta_{2}\mathbf v_{t-1}+(1-\beta_{2})\mathbf g_{t}^{\odot 2},\\
    \widehat{\mathbf m}_{t} &= \frac{\mathbf m_{t}}{1-\beta_{1}^{t}},
      & \widehat{\mathbf v}_{t} &= \frac{\mathbf v_{t}}{1-\beta_{2}^{t}},\\
    \mathbf\Theta_{t} &= \mathbf\Theta_{t-1}-\alpha\,\frac{\widehat{\mathbf m}_{t}}{\sqrt{\widehat{\mathbf v}_{t}}+\varepsilon}.
      & & \label{eq:adam_update}
\end{align}
Here \(\mathbf g_{t}^{\odot 2}\) denotes the componentwise square; the bias-corrected moments \(\widehat{\mathbf m}_{t}\) and \(\widehat{\mathbf v}_{t}\) compensate for the fact that \(\mathbf m_{t}\) and \(\mathbf v_{t}\) start at zero; and the division, square root, and addition of \(\varepsilon\) are all carried out componentwise. Here \(\alpha > 0\) is the base step size, \(\beta_{1},\beta_{2}\in[0,1)\) are decay parameters, and \(\varepsilon > 0\) serves for numerical stabilisation. Kingma and Ba recommend, as robust default values that perform well across a wide range of problems, the settings
    \(
    \alpha=10^{-3},
    \
    \beta_{1}=0.9,
    \
    \beta_{2}=0.999,
    \
    \varepsilon=10^{-8}.
    \)
    We retain these recommended values for \(\beta_{1}\), \(\beta_{2}\), and \(\varepsilon\); the experiments reported here, however, use the Nadam variant of Sect.~\ref{nadam} with a smaller initial step size of \(10^{-4}\) that is reduced on plateaus, as described in Sect.~\ref{numerical_experiments}.
    
    \begin{algorithm}[ht!]
        \floatname{algorithm}{Algorithm}
        \caption{\texttt{Adam} after Kingma and Ba \cite{kingma2014adam}. All vector operations are componentwise.}
        \label{alg:adam}
        \begin{algorithmic}[1]
            \Procedure{Adam}{}
            \Require Step size \(\alpha > 0\)
            \Require Decay parameters \(\beta_{1},\beta_{2}\in[0,1)\)
                \Require Stabilisation parameter \(\varepsilon > 0\)
                \Require Initial value \(\mathbf\Theta_{0}\)
                \State \(\mathbf m_{0}\gets \mathbf 0\)
                \State \(\mathbf v_{0}\gets \mathbf 0\)
                \State \(t\gets 0\)
                \While{stopping criterion not satisfied}
                \State \(t\gets t+1\)
                \State \(\mathbf g_{t}\gets \nabla_{\mathbf\Theta}\mathcal L_{t}(\mathbf\Theta_{t-1})\)
                \State \(\mathbf m_{t}\gets \beta_{1}\mathbf m_{t-1}+(1-\beta_{1})\mathbf g_{t}\)
                \State \(\mathbf v_{t}\gets \beta_{2}\mathbf v_{t-1}+(1-\beta_{2})\mathbf g_{t}^{\odot 2}\)
                \State \(\widehat{\mathbf m}_{t}\gets \mathbf m_{t}/(1-\beta_{1}^{t})\)
                \State \(\widehat{\mathbf v}_{t}\gets \mathbf v_{t}/(1-\beta_{2}^{t})\)
                \State \(\mathbf\Theta_{t}\gets \mathbf\Theta_{t-1}-\alpha\,\widehat{\mathbf m}_{t}/(\sqrt{\widehat{\mathbf v}_{t}}+\varepsilon)\)
                \EndWhile
                \State \Return \(\mathbf\Theta_{t}\)
                \EndProcedure
                \end{algorithmic}
                \end{algorithm}
                
                \texttt{Adam} is easy to implement, memory-efficient, and robust in many applications. In particular, the method is well suited to large parameter dimensions, noisy gradients, sparse gradients, and non-stationary objectives. The frequently encountered claim that \texttt{Adam} performs a kind of simulated annealing is, however, terminologically misleading. The method is not a simulated-annealing algorithm, but rather an adaptive first-order method with moment-based estimators. Likewise, \(\beta_{1}\) and \(\beta_{2}\) should not be understood as regularisation terms in the usual sense, but rather as the decay parameters of the exponential moving averages of the gradient statistics.
                
                For the present work, \texttt{Adam} is relevant above all because it is empirically very successful in deep convolution-based architectures and is highly compatible with mini-batch training. Variants such as \texttt{Adamax}, \texttt{AMSGrad}, or combinations with Nesterov momentum were proposed later, but they do not alter the basic idea of using adaptively scaled steps based on stochastic gradients and on their exponentially weighted first and second running moments.
                
                \subsection{Stochastic Gradient Descent with Nesterov-Adaptive Moment}
                \label{nadam}
                
                In order to improve a given deep-learning system, several strategies may be considered in principle, such as architectural modifications, more careful initialisation, or improved optimisation methods \cite{dozat2015incorporating,sutskever2013importance}. In the present setting, the choice of optimisation algorithm is particularly relevant, since the network parameters are updated in each iteration step on the basis of stochastic gradients. A classical device for accelerating gradient methods is the introduction of a momentum term \cite{polyak1964some}. The momentum term stores a smoothed average of earlier gradients, which damps oscillations across strongly curved directions of the loss surface while reinforcing those directions along which successive gradients agree~\cite{sutskever2013importance,dozat2015incorporating}.
                
                The classical momentum method augments each gradient step with a fixed fraction of the previously accumulated update direction, and in its simplest and most classical form reads
                \begin{align}
                    \mathbf g_t
                      & = 
                    \nabla_{\mathbf\Theta}\mathcal L_t(\mathbf\Theta_{t-1}),
                    \\
                    \mathbf m_t
                      & = 
                    \mu \mathbf m_{t-1} + \mathbf g_t,
                    \\
                    \mathbf\Theta_t
                      & = 
                    \mathbf\Theta_{t-1} - \eta \mathbf m_t,
                \end{align}
                where \(\mathbf\Theta_t\) denotes the parameter vector at step \(t\), \(\mathbf g_t\) the stochastic gradient, \(\mathbf m_t\) the momentum vector, \(\eta > 0\) the learning rate, and \(\mu \in [0,1)\) the momentum parameter.
                    
                    \begin{algorithm}[t]
                        \floatname{algorithm}{Algorithm}
                        \caption{Gradient method with momentum.}
                        \label{alg:gdmoment}
                        \begin{algorithmic}[1]
                            \State \(\mathbf g_t \gets \nabla_{\mathbf\Theta}\mathcal L_t(\mathbf\Theta_{t-1})\)
                            \State \(\mathbf m_t \gets \mu \mathbf m_{t-1} + \mathbf g_t\)
                            \State \(\mathbf\Theta_t \gets \mathbf\Theta_{t-1} - \eta \mathbf m_t\)
                        \end{algorithmic}
                    \end{algorithm}
                    
                    A drawback of classical momentum is that the gradient is evaluated at the current position \(\mathbf\Theta_{t-1}\), although the subsequent update already incorporates a movement in the direction of the old momentum \cite{sutskever2013importance}. This motivates Nesterov momentum, in which the gradient is evaluated at a forward-shifted position. In a common form, the method reads
                    \begin{align}
                        \mathbf g_t
                          & = 
                        \nabla_{\mathbf\Theta}\mathcal L_t\bigl(\mathbf\Theta_{t-1}-\mu \mathbf m_{t-1}\bigr),
                        \\
                        \mathbf m_t
                          & = 
                        \mu \mathbf m_{t-1} + \eta_t \mathbf g_t,
                        \\
                        \mathbf\Theta_t
                          & =
                        \mathbf\Theta_{t-1}-\mathbf m_t.
                    \end{align}
                    In this form the learning rate \(\eta_t\) is folded into the velocity \(\mathbf m_t\), whereas in the classical momentum method above it multiplies the velocity only in the parameter update; the two conventions are equivalent up to this rescaling. The idea is to probe the local geometry of the loss landscape already at a point in the direction to which the momentum term would move the method anyway.
                    Adam, recalled in \eqref{eq:adam_update}, replaces the plain momentum sum by an exponentially weighted average \(\widehat{\mathbf m}_t\) of past gradients and rescales each coordinate by the corresponding bias-corrected second-moment estimate \(\widehat{\mathbf v}_t\); its stochastic gradient \(\mathbf g_t=\nabla_{\mathbf\Theta}\mathcal L_t(\mathbf\Theta_{t-1})\) is, as before, evaluated on the mini-batch drawn at that step of the optimisation.
                    
                    Dozat proposes to incorporate the Nesterov idea into Adam by using not only the bias-corrected first moment \(\widehat{\mathbf m}_t\), but also a Nesterov-type look-ahead correction of the first moment \cite{dozat2015incorporating}. In Dozat's notation, let \(\mu_t\) be the time-dependent momentum parameter of the first moment and \(\nu\) the decay parameter of the second moment. The method computes the raw moments, forms a Nesterov-type look-ahead correction of the first moment, and finally takes an adaptively scaled step, according to the coupled parameter-update rules
\begin{align}
    \mathbf g_t &= \nabla_{\mathbf\Theta}\mathcal L_t(\mathbf\Theta_{t-1}),
      & \mathbf m_t &= \mu_t \mathbf m_{t-1} + (1-\mu_t)\mathbf g_t,\\
    \mathbf n_t &= \nu \mathbf n_{t-1} + (1-\nu)\mathbf g_t^{\odot 2},
      & \widehat{\mathbf n}_t &= \frac{\mathbf n_t}{1-\nu^t},\\
    \widehat{\mathbf m}_t &= \frac{\mu_{t+1}\mathbf m_t}{1-\prod_{i=1}^{t+1}\mu_i}
        + \frac{(1-\mu_t)\mathbf g_t}{1-\prod_{i=1}^{t}\mu_i},
      & \mathbf\Theta_t &= \mathbf\Theta_{t-1}-\alpha_t\,\frac{\widehat{\mathbf m}_t}{\sqrt{\widehat{\mathbf n}_t}+\epsilon}.
      \label{eq:nadam_update}
\end{align}
                    This is the usual form of the Nesterov-adaptive moment method. It combines the adaptive scaling of Adam with a Nesterov-type correction of the first moment \cite{dozat2015incorporating}.
                    
                    \begin{algorithm}[t]
                        \floatname{algorithm}{Algorithm}
                        \caption{Nadam after Dozat \cite{dozat2015incorporating}. All vector operations are componentwise.}
                        \label{alg:nadam}
                        \begin{algorithmic}[1]
                            \Procedure{Nadam}{}
                            \Require Step sizes \(\alpha_1,\dots,\alpha_T\)
                            \Require Momentum parameters \(\mu_1,\dots,\mu_{T+1}\)
                            \Require Second-moment parameter \(\nu \in [0,1)\)
                                \Require Stabilisation parameter \(\epsilon > 0\)
                                \Require Initial value \(\mathbf\Theta_0\)
                                \State \(\mathbf m_0 \gets \mathbf 0\)
                                \State \(\mathbf n_0 \gets \mathbf 0\)
                                \For{\(t=1,\dots,T\)}
                                \State \(\mathbf g_t \gets \nabla_{\mathbf\Theta}\mathcal L_t(\mathbf\Theta_{t-1})\)
                                \State \(\mathbf m_t \gets \mu_t \mathbf m_{t-1} + (1-\mu_t)\mathbf g_t\)
                                \State \(\mathbf n_t \gets \nu \mathbf n_{t-1} + (1-\nu)\mathbf g_t^{\odot 2}\)
                                \State \(\widehat{\mathbf m}_t \gets \mu_{t+1}\mathbf m_t / \bigl(1-\prod_{i=1}^{t+1}\mu_i\bigr) + (1-\mu_t)\mathbf g_t / \bigl(1-\prod_{i=1}^{t}\mu_i\bigr)\)
                                \State \(\widehat{\mathbf n}_t \gets \mathbf n_t / (1-\nu^t)\)
                                \State \(\mathbf\Theta_t \gets \mathbf\Theta_{t-1} - \alpha_t\,\widehat{\mathbf m}_t / \bigl(\sqrt{\widehat{\mathbf n}_t}+\epsilon\bigr)\)
                                \EndFor
                                \State \Return \(\mathbf\Theta_T\)
                                \EndProcedure
                                \end{algorithmic}
                                \end{algorithm}
                                
                                Nadam is therefore not a fundamentally new optimisation principle, but rather a targeted modification of Adam. Its practical benefit lies in an often somewhat faster and more stable optimisation, above all in deep networks whose stochastic gradients are noisy and whose loss surfaces are ill-conditioned. The exact superiority over Adam or classical stochastic gradient descent is, however, task-dependent and should not be formulated as a universal result. It is therefore more accurate, in the present work, to describe Nadam simply as the optimiser we selected for this particular experiment, rather than as a method that would be fundamentally superior across all learning problems and network architectures alike.
                                
                                \section{Numerical Experiments}
                                \label{numerical_experiments}
                                
                                The aim of the experiment is to learn a map from local tissue-density kernels to the corresponding dose voxel kernels. The starting point is a collection of precomputed density matrices that describe the local mass or material distribution in the neighbourhood of a voxel. For the same local configurations, the corresponding dose voxel kernels were computed by Monte Carlo simulation, as explained in Sect.~\ref{motivation}. The time-integrated activity distribution obtained from imaging can then be convolved with these density-adapted kernels in order to obtain the spatial distribution of the absorbed dose throughout the patient volume.
                                
                                The current standard approach uses fixed kernels for a small number of discrete tissue classes. In this approach, each voxel or local region is assigned to a class such as soft tissue, bone, or lung, and a separate dose voxel kernel is used for each class. The essential drawback of this procedure is that real anatomical structures often contain mixed tissue and continuous density transitions. A purely class-based discretisation of the CT information can approximate such continuous transitions only coarsely, and therefore leads, above all in heterogeneous regions, to systematic errors in the estimated dose at the boundaries between tissue types.
                                
                                The present experiment therefore investigates whether a neural network can predict the corresponding dose voxel kernel directly from a local density kernel. Formally, this is an image-to-image, or more precisely a volume-to-volume, regression problem. Since both the input and the target are spatially structured fields, we employ architectures that were originally developed for image analysis, image reconstruction, and segmentation. We first place the relevant related work in context and then define, precisely, the evaluation metrics by which the reconstruction quality is judged throughout the rest of this paper.
                                
                                \subsection{Related Work on Image Segmentation}
                                \label{related_work_on_image_segmentation}
                                
                                In image segmentation, the task is to assign a spatially resolved semantic structure to an image or volume, for example by assigning a class label to each pixel or voxel. Convolutional neural networks have had a major impact on this field. While early deep networks were used primarily for classification tasks, fully convolutional architectures demonstrated that the same basic ideas can also be used for dense prediction problems \cite{girshick2014rich,krizhevsky2017imagenet,ChenPKMY14,long2015fully}.
                                
                                Fully convolutional networks dispense with fully connected layers at the end of the architecture and instead produce a spatially structured output directly \cite{long2015fully}. As a consequence, they are suitable not only for segmentation, but also for depth estimation, image reconstruction, super-resolution, and more generally for image-to-image mappings \cite{long2015fully,eigen2013restoring,dong2014learning,liu2015deep}. This property makes them relevant for the present problem as well, since here one seeks to predict a spatial dose-response kernel from a purely local spatial input structure.
                                
                                A particularly influential architecture is the U-net, which was originally developed for biomedical image segmentation \cite{ronneberger2015u}. It consists of a contracting path for the extraction of increasingly abstract features and an expanding path for the reconstruction of a high-resolution output. Characteristic are the skip connections between layers of equal spatial resolution, which inject fine-grained localisation information from the contracting path directly into the expanding path. The volumetric extension of this idea was later formulated in the 3D U-net \cite{cciccek20163d}. Their favourable combination of local precision and global context makes U-net-type architectures the natural starting point for the volume-to-volume reconstruction task in this paper.
                                
                                \subsection{Evaluation Metrics for Reconstruction}
                                \label{measurements}
                                
                                To assess prediction quality, we use several complementary evaluation metrics. Since the data are represented as three-dimensional tensors of shape \((9,9,9)\), we write
                                \(
                                \mathbf X,\mathbf Y \in \mathbb R^{I\times J\times K},
                                \
                                1\le i\le I,\ 1\le j\le J,\ 1\le k\le K,
                                \)
                                and set
                                \(
                                N := IJK.
                                \)
                                In this notation \(\mathbf X\) denotes the target tensor produced by the Monte Carlo simulation, and \(\mathbf Y\) the corresponding tensor predicted by the trained network for that very same local density input, voxel by voxel.
                                
                                \subsubsection{Mean Absolute Error and Mean Squared Error}
                                
                                The mean squared error, our principal voxelwise measure of reconstruction accuracy for the regression task, penalises large deviations most heavily and is defined by
                                \begin{align}
                                    \operatorname{MSE}(\mathbf X,\mathbf Y)          
                                    =                                                
                                    \frac{1}{N}                                      
                                    \sum_{i=1}^{I}\sum_{j=1}^{J}\sum_{k=1}^{K}       
                                    \bigl(\mathbf Y_{ijk}-\mathbf X_{ijk}\bigr)^{2}. 
                                \end{align}
                                It penalises large pointwise errors more strongly than small ones and is therefore sensitive to outliers. It is a standard measure for regression problems, but it has no natural normalisation to a fixed interval and cannot be compared directly across different data sets.
                                
                                The mean absolute error, the second and complementary voxelwise measure that we use throughout the evaluation of the reconstructed dose voxel kernels, is given by
                                \begin{align}
                                    \operatorname{MAE}(\mathbf X,\mathbf Y)      
                                    =                                            
                                    \frac{1}{N}                                  
                                    \sum_{i=1}^{I}\sum_{j=1}^{J}\sum_{k=1}^{K}   
                                    \bigl|\mathbf Y_{ijk}-\mathbf X_{ijk}\bigr|. 
                                \end{align}
                                It is more robust to isolated large deviations than the quadratic error, and it is more directly interpretable. We use the two measures in a complementary way: although neither of them permits a standardised comparison across different data sets, together they give an immediate and interpretable description of the reconstruction error at each voxel of the kernel.
                                
                                Because the normalised targets and the network's sigmoid outputs both lie in \([0,1]\), every residual satisfies \(\mathbf Y_{ijk}-\mathbf X_{ijk}\in[-1,1]\), on which \(|x|^{2}\le|x|\). Summing this pointwise inequality over all voxels of the kernel yields the relation \(\operatorname{MSE}(\mathbf X,\mathbf Y)\le\operatorname{MAE}(\mathbf X,\mathbf Y)\), which serves as a convenient plausibility check on the implementation of both measures.
                                
                                \subsubsection{Intersection over Union}
                                
                                For two finite sets \(A,B\), the Jaccard coefficient, or intersection over union, is
                                \begin{align}
                                    \operatorname{Jaccard}(A,B)  
                                    =                            
                                    \frac{|A\cap B|}{|A\cup B|}. 
                                \end{align}
                                This quantity lies in the interval \([0,1]\) and measures the relative overlap of two sets.
                                
                                Since in the present problem one does not compare binary sets but rather nonnegative density tensors or dose voxel kernels, we replace intersection and union by their continuous counterparts, the voxelwise minimum and maximum, thereby defining the continuous overlap score, again taking values within the unit interval \([0,1]\), by the following expression
                                \begin{align}
                                    \operatorname{IoU}_{c}(\mathbf X,\mathbf Y)
                                    =                                               
                                    \frac{                                          
                                    \sum_{i=1}^{I}\sum_{j=1}^{J}\sum_{k=1}^{K}      
                                    \min\bigl(\mathbf X_{ijk},\mathbf Y_{ijk}\bigr) 
                                    }{                                              
                                    \sum_{i=1}^{I}\sum_{j=1}^{J}\sum_{k=1}^{K}      
                                    \max\bigl(\mathbf X_{ijk},\mathbf Y_{ijk}\bigr) 
                                    }.                                              
                                \end{align}
                                For nonnegative tensors with nonzero denominator, this quantity also takes values in \([0,1]\). It is equal to \(1\) exactly when \(\mathbf X=\mathbf Y\), and it becomes small when the two tensors differ substantially either spatially or in amplitude. Unlike the set-theoretic Jaccard coefficient, this is not literally an intersection-over-union measure for sets, but rather a continuous generalisation to nonnegative fields. For spatially localised energy deposition, this measure is useful because large contributions in the centre of a dose voxel kernel influence the value more strongly than the very small contributions near the boundary of the kernel.
                                
                                \subsubsection{Normalisation}
                                
                                Before training, the data are normalised componentwise by min--max scaling. Since the last layer of the network uses a sigmoid activation, it is convenient to map the target data as well into a subinterval of \((0,1)\). In order to avoid strong saturation of the sigmoid near \(0\) and \(1\), we do not scale the targets to the full interval \([0,1]\), but rather to the interior interval \([0.1,0.9]\).
                                
                                For \(a<b\) and a tensor \(\mathbf X\) with \(\max(\mathbf X)\neq \min(\mathbf X)\), the min--max scaling that maps an arbitrary entry \(\mathbf X_{ijk}\) of the tensor linearly onto the chosen target interval \([a,b]\) is given, entrywise, by
                                \begin{align}
                                    \operatorname{Norm}_{[a,b]}(\mathbf X_{ijk})                            
                                    =                                                                       
                                    (b-a)\,                                                                 
                                    \frac{\mathbf X_{ijk}-\min(\mathbf X)}{\max(\mathbf X)-\min(\mathbf X)} 
                                    +                                                                       
                                    a.                                                                      
                                \end{align}
                                Here \(\min(\mathbf X)\) and \(\max(\mathbf X)\) denote the minimum and maximum over all entries of the tensor \(\mathbf X\). In the degenerate case \(\max(\mathbf X)=\min(\mathbf X)\), this formula is undefined and must be handled separately, for example by assigning a constant value in the target interval.
                                
                                The restriction to an interior subinterval such as \([0.1,0.9]\) reduces the probability that the sigmoid function operates in nearly flat boundary regions. Keeping the sigmoid within its responsive range in this way stabilises the propagation of gradients through the output layer, and it does so without altering the relative order of the normalised data values.
                                
                                \subsection{Internal Covariate Shift}
                                \label{covariate_shift}
                                
                                Training deep convolutional neural networks is difficult because a large number of parameters must be optimised simultaneously. One of the difficulties emphasised by Ioffe and Szegedy is that, during training, the distribution of the inputs to a given layer changes whenever the parameters of earlier layers are updated \cite{ioffe2015batch}. They refer to this effect as \emph{internal covariate shift}. It was observed early on that centred and appropriately scaled data can substantially facilitate the training of neural networks \cite{lecun2012efficient,zeyer2017faster}. Batch normalisation transfers this idea into the network itself by normalising the layer inputs mini-batchwise during training.
                                
                                For a single layer that first computes an affine map of its input and then applies a pointwise nonlinearity to the result, a two-step composition that we may write compactly as
                                \begin{align}
                                    \mathbf y = \varphi(\mathbf W \mathbf x + \mathbf b), 
                                \end{align}
                                batch normalisation is, in its standard form, applied to the pre-activations \(\mathbf u = \mathbf W \mathbf x + \mathbf b\). Over a mini-batch \(\mathcal B=\{\mathbf u_{1},\dots,\mathbf u_{m}\}\) it first forms the componentwise batch mean and variance, then standardises every pre-activation by subtracting the mean and dividing by the batch standard deviation, and finally rescales the result by the learnable scale and shift parameters \(\boldsymbol{\gamma}\) and \(\boldsymbol{\beta}\), through the following coupled, componentwise assignments
                                \begin{align}
                                    \bar{\mathbf u}_{\mathcal B}
                                      & = 
                                    \frac{1}{m}\sum_{i=1}^{m}\mathbf u_i,
                                    \\
                                    \boldsymbol{\sigma}^{2}_{\mathcal B}
                                      & = 
                                    \frac{1}{m}\sum_{i=1}^{m}
                                    \bigl(\mathbf u_i-\bar{\mathbf u}_{\mathcal B}\bigr)^{\odot 2},
                                    \\
                                    \widehat{\mathbf u}_i
                                      & = 
                                    \frac{\mathbf u_i-\bar{\mathbf u}_{\mathcal B}}
                                    {\sqrt{\boldsymbol{\sigma}^{2}_{\mathcal B}+\epsilon}},
                                    \\
                                    \operatorname{BN}_{\boldsymbol{\gamma},\boldsymbol{\beta}}(\mathbf u_i)
                                      & = 
                                    \boldsymbol{\gamma}\odot \widehat{\mathbf u}_i+\boldsymbol{\beta},
                                \end{align}
                                Here \(\epsilon > 0\) guards against division by a vanishing batch variance, while the learnable parameters \(\boldsymbol{\gamma}\) and \(\boldsymbol{\beta}\) restore representational capacity, so that the full layer output becomes
                                \begin{align}
                                    \mathbf y_i                                                                                              
                                    =                                                                                                        
                                    \varphi\bigl(\operatorname{BN}_{\boldsymbol{\gamma},\boldsymbol{\beta}}(\mathbf W\mathbf x_i+\mathbf b)\bigr). 
                                    \label{eq:bn_standard}                                                                                   
                                \end{align}
                                
                                In the present experiment, normalisation is applied not before, but after, the activation function. This differs from the standard form \eqref{eq:bn_standard} and should therefore be stated explicitly as an architecture-specific design choice. In this case, the layer output is given by
                                \begin{align}
                                    \mathbf y_i                                                
                                    =                                                          
                                    \operatorname{BN}_{\boldsymbol{\gamma},\boldsymbol{\beta}} 
                                    \bigl(\varphi(\mathbf W\mathbf x_i+\mathbf b)\bigr).             
                                    \label{eq:bn_after_activation}                             
                                \end{align}
                                Such post-activation normalisation is possible, but it does not coincide with the original batch normalisation of Ioffe and Szegedy \cite{ioffe2015batch}. For convolution layers, the mean and variance are computed jointly over the mini-batch and over all spatial positions for each given channel, so that all activations produced by the same filter are normalised consistently \cite{ioffe2015batch}.
                                
                                For backpropagation through batch normalisation, the gradients are propagated through the affine rescaling, the standardisation, and the batchwise computation of mean and variance. Writing \(\tilde{\mathbf u}_i=\operatorname{BN}_{\boldsymbol{\gamma},\boldsymbol{\beta}}(\mathbf u_i)=\boldsymbol{\gamma}\odot\widehat{\mathbf u}_i+\boldsymbol{\beta}\) for the batch-normalised output, the required derivatives with respect to the scale and shift parameters \(\boldsymbol{\gamma}\) and \(\boldsymbol{\beta}\) are given explicitly by
                                \begin{align}
                                    \frac{\partial\mathcal L}{\partial \boldsymbol{\gamma}}
                                      = 
                                    \sum_{i=1}^{m}
                                    \frac{\partial\mathcal L}{\partial \tilde{\mathbf u}_i}
                                    \odot
                                    \widehat{\mathbf u}_i,
                                      \quad  
                                    \frac{\partial\mathcal L}{\partial \boldsymbol{\beta}}
                                       = 
                                    \sum_{i=1}^{m}
                                    \frac{\partial\mathcal L}{\partial \tilde{\mathbf u}_i},
                                \end{align}
                                while the derivative with respect to the normalised inputs is initially given by
                                \begin{align}
                                    \frac{\partial\mathcal L}{\partial \widehat{\mathbf u}_i} 
                                    =                                                         
                                    \frac{\partial\mathcal L}{\partial \tilde{\mathbf u}_i}           
                                    \odot                                                     
                                    \boldsymbol{\gamma}.                                      
                                \end{align}
                                The full backpropagation to \(\mathbf u_i\) is then obtained by a systematic application of the chain rule to the standardisation step \cite{ioffe2015batch}. Since these derivatives are implemented automatically in modern libraries, it suffices here to note that batch normalisation introduces additional learnable scale and shift parameters and can improve the numerical stability of training.
                                
                                \begin{algorithm}[t]
                                    \floatname{algorithm}{Algorithm}
                                    \caption{Batch normalisation for a mini-batch \cite{ioffe2015batch}.}
                                    \label{alg:ioffe2015batch}
                                    \begin{algorithmic}[1]
                                        \Procedure{Batch-Normalisation}{}
                                        \Require \(\mathcal B=\{\mathbf u_1,\dots,\mathbf u_m\}\)
                                        \Require \(\boldsymbol{\gamma},\boldsymbol{\beta}\)
                                        \State \(\bar{\mathbf u}_{\mathcal B}\gets \frac{1}{m}\sum_{i=1}^{m}\mathbf u_i\)
                                        \State \(\boldsymbol{\sigma}^{2}_{\mathcal B}\gets \frac{1}{m}\sum_{i=1}^{m}\bigl(\mathbf u_i-\bar{\mathbf u}_{\mathcal B}\bigr)^{\odot 2}\)
                                        \For{\(i=1,\dots,m\)}
                                        \State \(\widehat{\mathbf u}_i\gets (\mathbf u_i-\bar{\mathbf u}_{\mathcal B})/\sqrt{\boldsymbol{\sigma}^{2}_{\mathcal B}+\epsilon}\)
                                        \State \(\tilde{\mathbf u}_i\gets \boldsymbol{\gamma}\odot \widehat{\mathbf u}_i+\boldsymbol{\beta}\)
                                        \EndFor
                                        \State \Return \(\tilde{\mathbf u}_1,\dots,\tilde{\mathbf u}_m\)
                                        \EndProcedure
                                    \end{algorithmic}
                                \end{algorithm}
                                
                                \subsection{Residual Networks}
                                \label{residual_neural_networks}
                                
                                Residual networks are based on the idea of learning a residual map rather than a direct target map \cite{he2016identity}. If \(\mathbf x_n\) is the input of a residual block and \(\mathcal F(\mathbf x_n\mid \mathbf W_n)\) is the residual function realised by the weighted layers of the block, then the standard form of the block is
                                \begin{align}
                                    \mathbf y_n
                                      & = 
                                    \mathbf x_n + \mathcal F(\mathbf x_n\mid \mathbf W_n),
                                    \\
                                    \mathbf x_{n+1}
                                      & = 
                                    \varphi(\mathbf y_n).
                                \end{align}
                                In the case of a pre-activation architecture, the activation may also be placed inside the block. What is essential is that the skip connection combines the input additively with the residual function \cite{he2016identity}. Concatenation, by contrast, is a different architectural idea, as used for example in U-net-type skip connections, but it is not the classical residual block.
                                
                                If the skip connection is the identity and, in addition, the after-block activation \(\varphi\) is the identity, as in the pre-activation residual block, then repeated substitution of this block relation unrolls the hidden state of an arbitrarily deep layer of the network into the closed additive form
                                \begin{align}
                                    \mathbf x_{N}                                            
                                    =                                                        
                                    \mathbf x_{n}                                            
                                    +                                                        
                                    \sum_{i=n}^{N-1}\mathcal F(\mathbf x_i\mid \mathbf W_i). 
                                \end{align}
                                It follows immediately that the backward gradient flow splits into a direct term through the identity connection and one through the residual branches. Formally, differentiating this identity with respect to \(\mathbf x_n\) by the chain rule yields the additive decomposition
                                \begin{align}
                                    \frac{\partial \mathcal L}{\partial \mathbf x_n}        
                                    =                                                       
                                    \frac{\partial \mathcal L}{\partial \mathbf x_N}        
                                    \left(                                                  
                                    \mathbf I                                               
                                    +                                                       
                                    \frac{\partial}{\partial \mathbf x_n}                   
                                    \sum_{i=n}^{N-1}\mathcal F(\mathbf x_i\mid \mathbf W_i) 
                                    \right),                                                
                                \end{align}
                                where \(\mathbf I\) denotes the identity map. This direct additive term is the reason why residual connections facilitate gradient flow and make very deep networks trainable \cite{he2016identity}.
                                
                                In the present experiment, residual connections are used within the U-net-type architecture in order to keep locally learned features accessible even across deeper parts of the network. One should, however, distinguish conceptually between \emph{residual addition} and \emph{U-net concatenation}. The former adds two paths of equal dimension and leaves the channel count unchanged, whereas the latter stacks the feature maps of both paths along the channel axis.
                                
                                \begin{figure}[htbp]
                                    \centering
  \adjustbox{max width=\linewidth}{\footnotesize\color{burgundydark}
\begin{tikzcd}[column sep=1.6em, row sep=2.4em]
  \mathbf{W}_{n+1} \arrow[r] & \text{ReLU} \arrow[r] & \text{BN} \arrow[r] & \mathbf{W}_{n+2} \arrow[r] & \text{ReLU} \arrow[r] & \text{BN} \arrow[d] & \\
  \mathbf{x}_n \arrow[u] \arrow[rrrrr, gold, "\text{identity}" description] & & & & & {\oplus} \arrow[r] & \mathbf{y}_n
\end{tikzcd}}
                                    \caption{Schematic representation of a residual block. The skip connection is combined additively with the residual branch, and batch normalisation and the activations may be placed before or after the weighted transformation.}
                                \end{figure}
                                
                                \subsection{Dropout}
                                \label{sec:srivastava2014dropout}
                                
                                Dropout is a simple and effective regularisation technique for reducing overfitting in neural networks \cite{srivastava2014dropout}. During training, randomly selected units in a layer, together with their outgoing connections, are deactivated. This prevents the network from relying too heavily on specific co-activations of individual neurons and instead encourages it to form more robust and more redundant internal representations of its high-dimensional input patterns.
                                
                                Let \(\mathbf y_{n-1}\) be the activation of the \((n-1)\)-st layer. Without any dropout applied, forward propagation in the \(n\)-th layer proceeds in the usual two steps and has the form
                                \begin{align}
                                    \mathbf z_n
                                      = 
                                    \mathbf W_n \mathbf y_{n-1} + \mathbf b_n,
                                    \quad
                                    \mathbf y_n
                                      = 
                                    \varphi(\mathbf z_n).
                                \end{align}
                                With dropout, one first draws, independently for the individual units of the layer, a random binary retention mask whose entries are independently distributed according to
                                \begin{align}
                                    \mathbf r_{n-1}             
                                    \sim                        
                                    \operatorname{Bernoulli}(p) 
                                \end{align}
                                componentwise, where \(p\in[0,1]\) is the keep probability. One then sets
                                \begin{align}
                                    \widetilde{\mathbf y}_{n-1}           
                                    =                                     
                                    \mathbf r_{n-1}\odot \mathbf y_{n-1}, 
                                \end{align}
                                and then uses this masked activation \(\widetilde{\mathbf y}_{n-1}\) in place of the original activation \(\mathbf y_{n-1}\):
                                \begin{align}
                                    \mathbf z_n
                                      = 
                                    \mathbf W_n \widetilde{\mathbf y}_{n-1} + \mathbf b_n,
                                    \quad
                                    \mathbf y_n
                                      = 
                                    \varphi(\mathbf z_n).
                                \end{align}
                                In modern implementations, one usually uses inverted dropout. In that variant, one already rescales the retained activations by the factor \(1/p\) during training itself:
                                \begin{align}
                                    \widetilde{\mathbf y}_{n-1}                        
                                    =                                                  
                                    \frac{1}{p}\,\mathbf r_{n-1}\odot \mathbf y_{n-1}, 
                                \end{align}
                                so that no additional scaling is required at test time \cite{srivastava2014dropout}.
                                
                                The often-quoted interpretation that a network with \(n\) units behaves like an ensemble of \(2^n\) thinned subnetworks is heuristically useful, but should not be misunderstood as an exact structural decomposition \cite{srivastava2014dropout}. For the mathematical description, it is enough to observe that dropout implements stochastic regularisation by random masking. In the present experiment, dropout is used to improve the generalisation performance of the network under limited data.
                                
                                \begin{figure}[t]
                                    \centering
  \adjustbox{max width=\linewidth}{\footnotesize\color{burgundydark}
\begin{tikzpicture}[>=Stealth, font=\small,
  blk/.style={draw=burgundy, line width=0.8pt, fill=burgundy!9, rounded corners=1.2pt,
              minimum height=13mm},
  cube/.style={draw=burgundydark, line width=0.8pt, fill=burgundywash, minimum size=9mm},
  flow/.style={draw=burgundydark, line width=0.8pt, ->},
  skip/.style={draw=gold, line width=0.8pt, ->, densely dashed},
  res/.style ={circle, draw=burgundydark, thick, fill=white, inner sep=0.6pt,
               font=\scriptsize}]

  \node[cube] (in)  at (0,0) {};
  \node[font=\scriptsize] at (0,-1.15) {density};
  \node[font=\scriptsize] at (0,-1.55) {$9^3$};

  \node[blk, minimum width=3.0mm] (e1) at (1.7,0) {};
  \node[blk, minimum width=4.5mm] (e2) at (3.0,0) {};
  \node[blk, minimum width=6.0mm] (e3) at (4.4,0) {};
  \node[blk, minimum width=7.2mm] (m)  at (6.0,0) {};
  \node[blk, minimum width=6.0mm] (d3) at (7.6,0) {};
  \node[blk, minimum width=4.5mm] (d2) at (9.0,0) {};
  \node[blk, minimum width=3.0mm] (d1) at (10.3,0){};
  \node[cube] (out) at (12.0,0) {};
  \node[font=\scriptsize] at (12.0,-1.15) {dose};
  \node[font=\scriptsize] at (12.0,-1.55) {$9^3$};

  \draw[flow] (in) -- (e1);
  \draw[flow] (e1) -- (e2);
  \draw[flow] (e2) -- (e3);
  \draw[flow] (e3) -- (m);
  \draw[flow] (m)  -- (d3);
  \draw[flow] (d3) -- (d2);
  \draw[flow] (d2) -- (d1);
  \draw[flow] (d1) -- (out);

  \draw[skip] (e3.north) to[bend left=28] (d3.north);
  \draw[skip] (e2.north) to[bend left=24] (d2.north);
  \draw[skip] (e1.north) to[bend left=20] (d1.north);

  \node[res] at ([yshift=-1.6mm]e2.south) {$+$};
  \node[res] at ([yshift=-1.6mm]e3.south) {$+$};
  \node[res] at ([yshift=-1.6mm]m.south)  {$+$};
  \node[res] at ([yshift=-1.6mm]d3.south) {$+$};
  \node[res] at ([yshift=-1.6mm]d2.south) {$+$};

  \begin{scope}[shift={(0.2,-2.5)}]
    \draw[flow] (0,0) -- (0.7,0);
    \node[anchor=west, font=\scriptsize] at (0.8,0) {conv $+$ BN $+$ LeakyReLU};
    \draw[skip] (5.3,0) -- (6.0,0);
    \node[anchor=west, font=\scriptsize] at (6.1,0) {concatenation (skip)};
    \node[res] (rr) at (9.3,0) {$+$};
    \node[anchor=west, font=\scriptsize] at (9.55,0) {residual addition};
  \end{scope}
\end{tikzpicture}%
}
                                    \caption{Architecture of the U-net-type model for estimating dose voxel kernels. All convolution layers except the last are combined with \texttt{LeakyReLU} and batch normalisation. The lateral U-net connections are implemented as concatenations of feature maps of equal spatial resolution.}
                                    \label{abb:dvk_nn_architecture}
                                \end{figure}
                                
                                \subsection{Learning Density-Dependent Dose Voxel Kernels by Means of U-Residual Networks}
                                \label{sec:learningwithunets}
                                
                                \definecolor{hnr}{RGB}{176,0,52}
                                
                                To predict density-dependent dose voxel kernels, we use a U-net-type architecture with additional residual connections. The choice of this model class is motivated by the fact that the task is a volume-to-volume regression problem: a local dose voxel kernel is to be predicted from a local density kernel. U-net architectures combine a contracting path for feature extraction with an expanding path for the reconstruction of a spatially resolved output. The lateral skip connections transfer localisation-relevant features directly between levels of equal spatial resolution. In the present model, the spatial resolution is preserved to a large extent, so that the architecture is better understood as a shallow hierarchical U-residual network without aggressive dimensional reduction. In particular, no strong downsampling is used, since the precise spatial structure of the dose voxel kernel is to be retained.
                                
                                The implementation was carried out in \texttt{Keras v2.1.2} \cite{chollet2015keras} using \texttt{TensorFlow v1.5} \cite{tensorflow2015-whitepaper}. As optimisation method, we used \texttt{NADAM}, the Nesterov-adaptive moment method; see Sect.~\ref{nadam} for its precise definition, including the full set of update equations.
                                
                                \subsubsection{Neural Network Architecture}
                                
                                The network used comprises \(45\) layers in total. The convolution kernels were initialised using a LeCun-type initialisation \cite{lecun2012efficient}. Each convolution is followed by a \texttt{LeakyReLU} activation and a batch-normalisation layer. Since the exact negative slope of the \texttt{LeakyReLU} is not essential for the mathematical description of the architecture, we refer to it below simply as a fixed parameter that is not learned during training. All convolution layers were regularised. The first convolution layer carried an \(L_{1}\) penalty of coefficient \(\lambda=0.005\) on its weights, whereas every subsequent convolution layer instead carried a weaker \(L_{2}\) penalty of coefficient \(\lambda=0.001\), so that the regularisation acts most strongly at the input layer of the network.
                                
                                \subsubsection{Data and Tissue Classes}
                                
                                The data set consists of equal proportions of mass-density kernels and corresponding dose voxel kernels from the tissue classes lung, kidney, liver, bone, and spleen. For each tissue class, \(2000\) pairs of density and dose data are available, giving a total of \(10000\) samples. Since the tissue classes differ substantially in their structure, the data were shuffled randomly before training, and the mini-batches were then drawn randomly from this mixed data set. The data set was split into \(7000\) training examples and \(3000\) validation examples. The validation set was held completely separate from the optimisation process during training, and the mini-batch drawn at each optimisation step had a fixed size of exactly \(128\) examples.
                                
                                \begin{figure}[!ht]
    \centering
    \includegraphics[width=\textwidth]{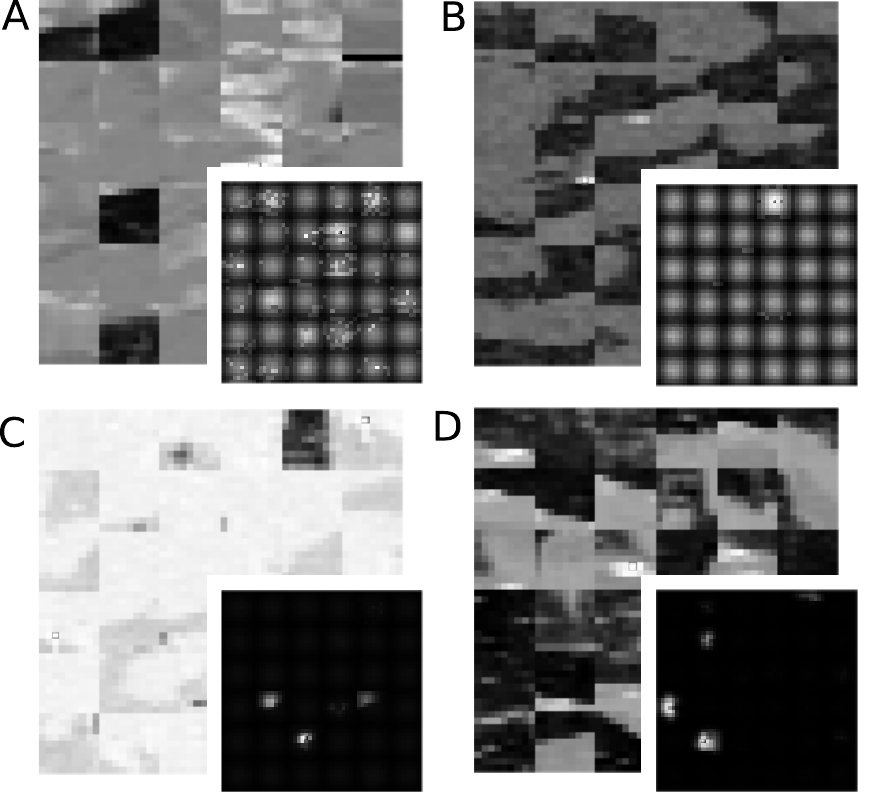}
    \caption{Cross-sections of mass-density kernels and the corresponding dose
    voxel kernels along the \(i\)-th axis. The larger panels show the mass
    density of a local tissue kernel and the smaller panels the corresponding
    absorbed dose; matching panels form a pair, each drawn independently and at
    random from the data set. Example A is bone, B kidney, C liver, D lung, and
    E spleen tissue, with \(36\) cross-sectional pairs per tissue type. In the
    density panels darker means denser; in the dose panels brighter means a
    higher deposited dose.}
    \label{fig:cross_sections}
\end{figure}

\begin{table}[!ht]
    \centering
    \caption{Reconstruction quality by tissue class after \(308\) training
    epochs, on the training split and on the held-out validation split. The
    intersection-over-union overlap is dimensionless in \([0,1]\); the mean
    absolute error is reported in units of \(10^{-2}\) and the mean squared
    error in units of \(10^{-4}\), both on the min--max--normalised kernels. The
    \emph{total} row is evaluated over the pooled validation set and is not the
    arithmetic mean of the per-class rows; the best single tissue class is
    lung tissue, and the overall results are set in bold.}
    \label{tab:results}
    \begin{tabular}{lcccccc}
        \toprule
        & \multicolumn{2}{c}{\textbf{IoU} \([0,1]\)}
        & \multicolumn{2}{c}{\textbf{MAE} \(\times 10^{-2}\)}
        & \multicolumn{2}{c}{\textbf{MSE} \(\times 10^{-4}\)}\\
        \cmidrule(lr){2-3}\cmidrule(lr){4-5}\cmidrule(lr){6-7}
        \textbf{Tissue} & train & test & train & test & train & test\\
        \midrule
        bones  & $0.55$ & $0.47$ & $3.91$ & $4.23$ & $1.18$ & $1.21$\\
        lungs  & $0.94$ & $0.90$ & $0.12$ & $0.19$ & $0.78$ & $0.97$\\
        kidney & $0.73$ & $0.72$ & $3.10$ & $3.30$ & $1.12$ & $1.82$\\
        liver  & $0.82$ & $0.79$ & $2.41$ & $2.78$ & $1.00$ & $1.26$\\
        spleen & $0.64$ & $0.61$ & $4.05$ & $5.01$ & $1.68$ & $1.81$\\
        \midrule
        \textbf{total} & $\mathbf{0.96}$ & $\mathbf{0.86}$ & $\mathbf{2.29}$ & $\mathbf{2.12}$ & $\mathbf{1.18}$ & $\mathbf{1.24}$\\
        \bottomrule
    \end{tabular}
\end{table}

\begin{figure}[!ht]
    \centering
    \includegraphics[width=\textwidth]{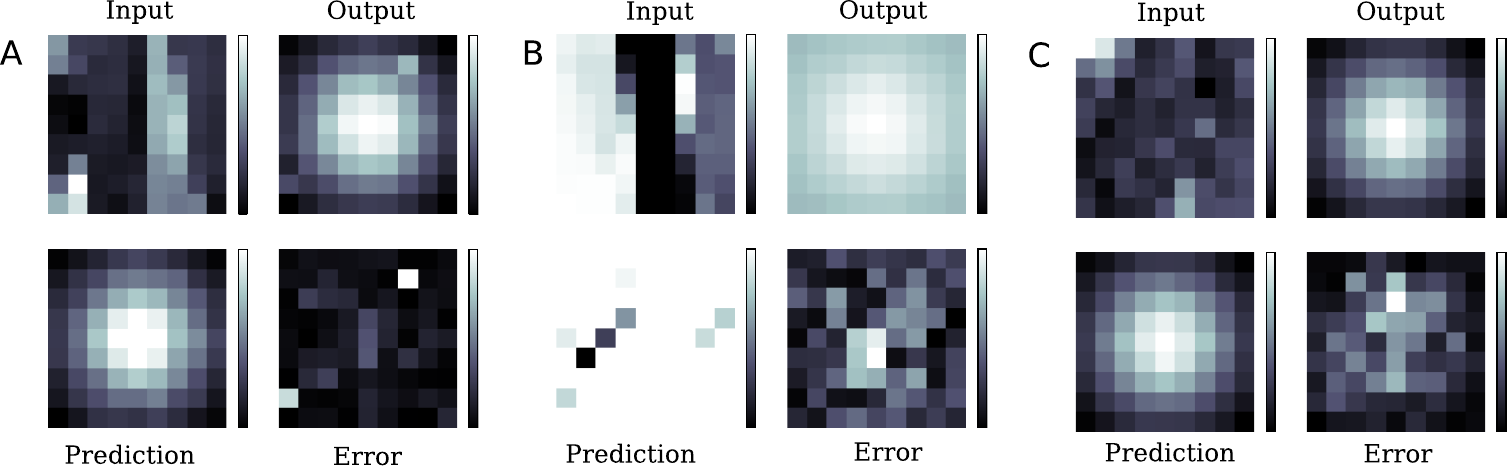}
    \caption{Predicted versus target dose voxel kernels. Cross-section A is from
    liver, B from kidney, and C from bone tissue. In each case the slice is the
    fifth of the nine sections along the first spatial axis, chosen because most
    of the activity, and hence of the simulated source, sits at the centre of
    the density kernel. The predicted amplitudes closely match those of the
    Monte Carlo targets, and in examples A and C the spatial pattern of the
    deposited energy is reconstructed faithfully down to fine structural detail
    within the kernel volume.}
    \label{abb:densities}
\end{figure}

                                \subsubsection{Experimental Results}
                                
\begin{figure}[!ht]
  \centering
  \includegraphics[width=\linewidth]{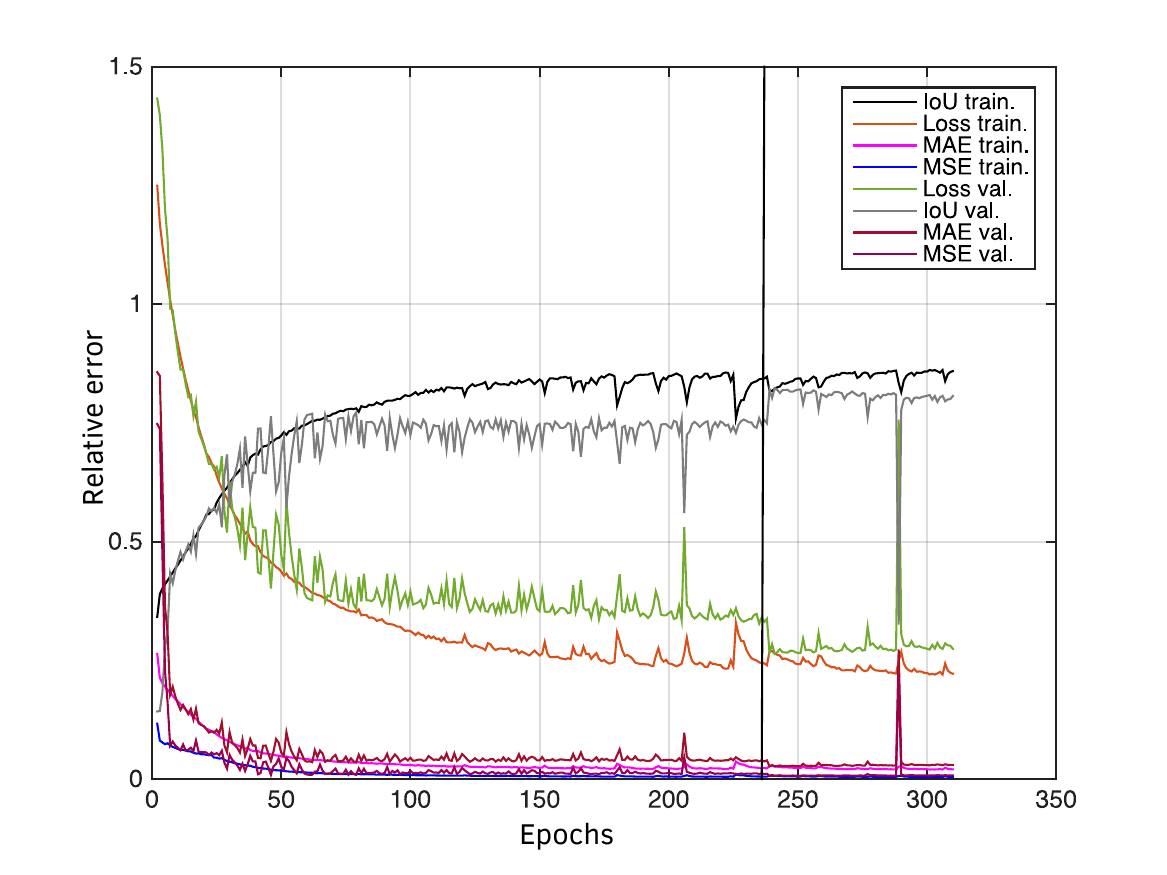}
  \caption{Training and validation curves for the mean squared error, the mean
  absolute error, and the continuous intersection-over-union overlap score, on
  the \(7:3\) training-to-validation split. The initial learning rate was
  \(10^{-4}\); the vertical line at epoch \(262\) marks where it was halved to
  \(0.5\times10^{-4}\). The training objective was the complement
  \(1-\mathrm{IoU}\) of the overlap score, so that minimising the loss maximises
  overlap. The learning rate was reduced on this same plateau criterion, while
  training as a whole ran for \(308\) epochs, at which point early stopping on the
  validation overlap terminated the training run.}
  \label{fig:training_curves}
\end{figure}

                                On the held-out validation split the network generalised well. Training minimised the complement \(1-\mathrm{IoU}\) of the continuous overlap score, so that decreasing the loss increases overlap; after \(308\) epochs the overlap score reached \(0.86\), and the learning curves are shown in Fig.~\ref{fig:training_curves}. Optimisation started from a learning rate of \(10^{-4}\), which was halved whenever the loss failed to improve by at least \(\epsilon=10^{-6}\) over \(15\) consecutive epochs. The inputs were min--max--normalised mass-density kernels and the targets the correspondingly normalised dose voxel kernels. Table~\ref{tab:results} reports the error metrics by tissue class, whose \emph{total} row aggregates over the pooled validation set rather than averaging the classes; the strongest single class is lung tissue. Representative density and dose kernels appear in Fig.~\ref{fig:cross_sections}, and predicted kernels against their Monte Carlo targets in Fig.~\ref{abb:densities}, where their agreement is directly visible.
                                
                                These results show that the proposed architecture is capable of approximating a nontrivial map from local density structures to local dose responses as determined by Monte Carlo simulation. This claim should be stated deliberately cautiously. The model does not replace the underlying transport physics; it learns only a fast, data-driven approximation of the map that the Monte Carlo simulation itself computes. The strong performance on the independent validation split indicates that the model has captured the underlying density-to-dose map, rather than simply memorising the particular training examples it was shown during fitting.
                                
                                In the course of data analysis, principal-component methods for dimensionality reduction were also examined. It turned out, however, that the variance was not concentrated in a way that would have justified a strongly dimension-reducing projection without significant loss of information. For this reason, no preliminary dimensionality reduction was applied.
                                
\begin{figure}[!ht]
  \centering
  \includegraphics[width=\linewidth]{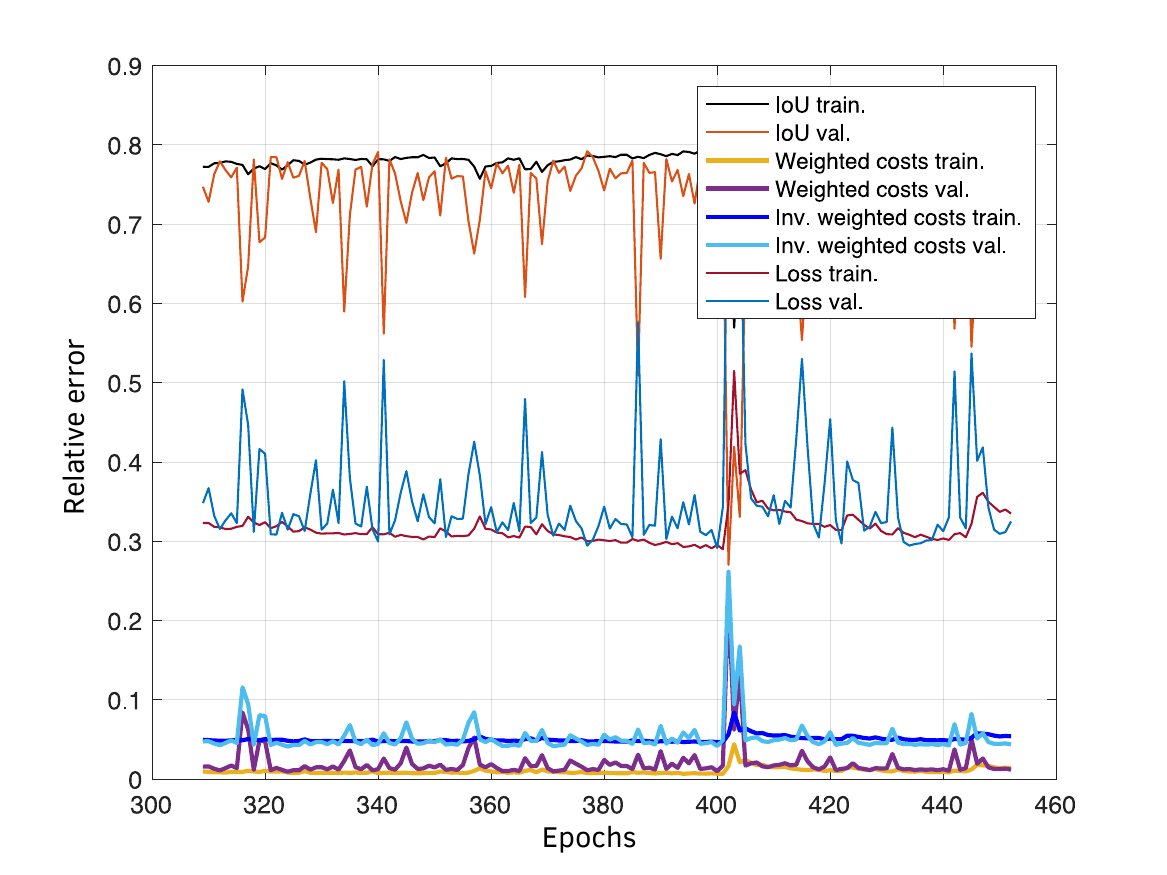}
  \caption{Evaluation under the clinically motivated, energy-weighted
  objective~\eqref{costs_clinic}. The energy-weighted curves remain strongly correlated
  with the corresponding unweighted metrics, so that the weighting sharpens the
  emphasis on high-dose regions without distorting the overall trend; this run
  was continued to epoch~\(453\) to confirm the stability of the behaviour.}
  \label{fig:clinical_loss}
\end{figure}

                                \section{Discussion}
                                \label{discussion}
                                
                                The experiments show that deep learning architectures are, in principle, suitable for approximating density-dependent maps arising in dosimetry. The method developed here learns, from density kernels, local weights of a spatial response structure that may be interpreted as an approximation of the dose voxel kernels obtained from Monte Carlo simulation. The model can therefore also be applied to data that were not seen during training. The error metrics provide a first quantitative assessment of accuracy, but they capture the clinical or physical relevance of a given reconstruction error only partially, as discussed next.
                                
                                In particular, one must note that an error at the centre of a dose voxel kernel may be clinically much more serious than an error in the periphery. The reason is that a substantial part of the deposited energy is concentrated in the immediate vicinity of the source voxel. A purely geometric or pointwise equally weighted error function does not reflect this asymmetry adequately. It is therefore natural to weight the error physically, letting voxels that carry a high deposited target energy contribute more strongly to the total than peripheral voxels, in which comparatively little energy is deposited and an error is correspondingly less important.
                                
                                A natural choice is a squared error weighted by the relative target energy. Keeping the convention of Sect.~\ref{measurements}, in which \(\mathbf X\) is the Monte Carlo target kernel and \(\mathbf Y\) the network output, a physically well-motivated loss function is the target-energy-weighted squared error
                                \begin{align}
                                    \mathcal L_{\mathrm{phys}}(\mathbf X,\mathbf Y)      
                                    =                                                    
                                    \sum_{i,j,k}                                         
                                    \bigl(\mathbf Y_{ijk}-\mathbf X_{ijk}\bigr)^{2}
                                    \frac{\mathbf X_{ijk}}{\sum_{i,j,k}\mathbf X_{ijk}},
                                    \label{costs_clinic}                                 
                                \end{align}
                                provided that \(\sum_{i,j,k}\mathbf X_{ijk} > 0\). This function is a squared error with normalised, target-dependent weights. Positions of high physical relevance thus contribute more strongly to the total error than peripheral regions with low energy deposition. The quantity defined in \eqref{costs_clinic} is therefore not a mean squared error in the strict sense, but a squared error in which each voxel is weighted by its own share of the total deposited target energy.
                                
                                The present results are encouraging, but they are not yet sufficient for a clinical assessment. In future investigations, full patient-specific dose calculations will need to be compared with the predictions produced by the neural networks. In particular, it will be necessary to examine to what extent the method reduces systematic deviations in relevant anatomical regions relative both to Monte Carlo simulation and to currently used standard methods. Only on this basis can one determine whether the method is suitable for clinical applications.
                                
                                \paragraph{Acknowledgements}
                                
                                I thank Bernd Ludwig and Elmar Lang for their supervision, and Martin Böddecker for providing the hardware used for the experiments. I also thank Dominique Melodia, Beata Melodia, Domenico Melodia, and Marie-Louise Isenberg for proofreading and support in the preparation of this work. I thank Sebastian Müller, Thomas Büttner, Tobias Baron, and Philipp Gäbelein for stimulating technical discussions. The data were kindly provided by University Hospital Erlangen, whose support is gratefully acknowledged.
                                
                                \paragraph{Code}
                                
                                The source code for the experiments is available at the following address:
                                \url{https://codeberg.org/Jiren/MADVK}. The repository also contains the training and evaluation scripts required to reproduce all of the numerical results reported in this paper.
                                \bibliographystyle{plainnat}
                                \bibliography{biblio}
\end{document}